\documentclass[10pt,twocolumn,letterpaper]{article}
\usepackage[pagenumbers]{cvpr}

\usepackage{amsmath,amsfonts,bm}

\def\eqref#1{equation~\ref{#1}}

\def\1{\bm{1}}

\DeclareMathAlphabet{\mathsfit}{\encodingdefault}{\sfdefault}{m}{sl}
\SetMathAlphabet{\mathsfit}{bold}{\encodingdefault}{\sfdefault}{bx}{n}

\usepackage{float}
\usepackage{microtype}
\usepackage{subcaption}
\usepackage{multirow}
\usepackage{array}
\usepackage{placeins}
\newcommand{\cmark}{\checkmark}
\newcommand{\xmark}{$\times$}
\definecolor{cvprblue}{rgb}{0.21,0.49,0.74}
\usepackage[pagebackref,breaklinks,colorlinks,allcolors=cvprblue]{hyperref}
\def\confName{CVPR}
\def\confYear{2027}
\title{MIBE: Multi-subject Interaction Benchmark and Evaluator\\for Personalized Image Generation}
\author{%
    Zhihan Chen\thanks{Equal contribution}\\
    University of California, Los Angeles\\
    \texttt{chenz23@ucla.edu}\\
    \and
    Yuhuan Zhao\footnotemark[1]\\
    University of Southern California\\
    \texttt{yuhuanzh@usc.edu}\\
    \and
    Yijie Zhu\footnotemark[1]\\
    DeerLab LLC\\
    \texttt{ ejzhu2025@gmail.com}\\
    \and
    Xinyu Yao\footnotemark[1] \\
    Carnegie Mellon University\\
\texttt{xinyuyao@andrew.cmu.edu} \\
    \and
    Mengcong Ren\\
    Clemson University\\
\texttt{mengcongren@gmail.com}\\
    \and
    Suwen Wang\\
    University of California, Los Angeles\\
    \texttt{suwenw@g.ucla.edu}\\
    \and 
    Qiuyang Yin\\
    Independent Researcher\\
\texttt{qiuyang.yin.16@gmail.com}
    \and
    Yuchen Sun\\
    San Jose State University\\
\texttt{yuchen.sun01@sjsu.edu}\\
    \and
    Qin Wang\\
    University of Illinois at Urbana-Champaign\\
    \texttt{qw23@illinois.edu}\\
    \and 
    Lu Xin\\
    DeerLab LLC\\
\texttt{lucasxinlu@outlook.com}
}
\begin{document}
\maketitle
\begin{abstract}
Multi-subject personalized image generation aims to compose multiple distinct reference identities into a coherent scene while preserving their prompt-specified interactions. Existing generators frequently exhibit \emph{subject-binding failures} as scene complexity scales: omitting requested entities (\emph{Existence}), blending visual attributes across subjects (\emph{Appearance}), or misassigning actions and spatial relations (\emph{Interaction}). Compounding this modeling bottleneck, existing evaluation metrics, which are designed primarily for single-subject similarity or holistic image preference, align only weakly with human judgment on these coupled errors, and several degrade toward random agreement as the number of interacting subjects increases. To address this evaluation gap, we introduce the Multi-subject Interaction Benchmark and Evaluator (MIBE), a unified framework pairing a factorized benchmark (MIB) with a reference-conditioned evaluator (MIE). MIB systematically varies subject count, human-heavy/object-heavy composition, and relation type (non-contact, occlusion, and physical interaction) within a decoupled two-tier data regime: MIB-Silver, a 60K-pair training corpus annotated via a rubric-guided dual-VLM protocol, and MIB-Gold, a 4{,}020-pair evaluation suite (3{,}690 pairs with a consensus winner) annotated under a double-blind human protocol across six state-of-the-art generators spanning both seen and held-out unseen architectures. Trained exclusively on MIB-Silver with a dual-head objective that jointly predicts pairwise preference rankings and facet-level diagnostics, MIE achieves 0.922 overall pairwise agreement with human consensus on MIB-Gold, including 0.982 on seen generators and 0.884 on held-out unseen generators. Using MIBE for cross-generator diagnosis, we further quantify systematic binding bottlenecks across modern generators, providing both reproducible benchmark infrastructure and interpretable supervision for future model alignment.
\end{abstract}

\section{Introduction}

Personalized text-to-image generation aims to synthesize specific reference subjects within new scenes described by a text prompt. While existing methods are able to preserve identity for single-subject cases~\citep{gal2022image,ruiz2023dreambooth,li2024photomaker}, composing two or more distinct reference subjects that interact within the same scene remains challenging~\citep{xiao2025fastcomposer,wang2025ms,she2026mosaic,chen2025xverse}. In such multi-subject settings, the generator must address what we term the \emph{multi-subject binding} problem across three criteria: (1)~every requested reference entity must appear as a separate physical instance without omission or duplication (\emph{Existence}); (2)~each instance must preserve the facial identity, structure, and attire of its own reference without cross-subject attribute leakage (\emph{Appearance}); and (3)~the subjects must execute the spatial relations, occlusions, and physical actions specified by the prompt (\emph{Interaction}). These three dimensions are structurally coupled in generative outputs rather than independent: when a generator omits a requested person, it can graft that missing person's suit, tie, or facial traits onto another surviving figure in the scene, or duplicate a single identity to satisfy two concurrent actions separately (Appendix~\ref{sec:appendix_failure_modes}, Figures~\ref{fig:appendix-failure-rates-subject-count}--\ref{fig:appendix-multi-action-overload}). Such binding failures are especially severe in contact-rich and occlusion-heavy scenes, where overlapping boundaries make local instance attribution ambiguous.

\begin{figure}[!ht]
  \centering
  \includegraphics[width=1.0\linewidth]{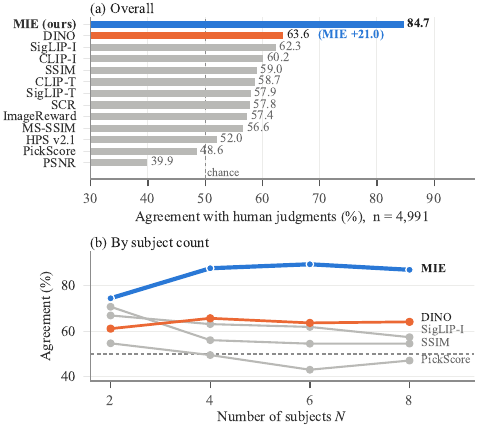}
    \caption{
      Similarity- and preference-based metrics align weakly with human preference; MIE aligns substantially better.
      All methods are scored on the same 4{,}991 individual human judgments of the unseen-generator split of MIB-Gold, with ties counted as half agreement. (a) Overall agreement; the strongest metric (DINO) is highlighted. (b) Agreement by subject count ($N \in \{2,4,6,8\}$) for MIE, DINO, and three representative metrics. 
    }
  \label{fig:metric_agreement_subject_count}
\end{figure}

Existing automatic metrics provide incomplete measures of these cross-subject binding failures. Reference-similarity metrics~\citep{radford2021learning,zhai2023sigmoid,oquab2023dinov2} and face-matching scores~\citep{deng2019arcface,borse2025multihuman} measure feature overlap between references and generated crops, but they do not verify which identity performs which action, and can therefore reward a missing subject's clothing transferred onto the wrong person. Conversely, general-purpose preference reward models~\citep{kirstain2023pick,xu2023imagereward,wu2023human} judge generations without conditioning on the reference images, and so cannot check whether each referenced subject is present and correctly bound. Measured against double-blind human judgments on scenes with $N=2$ to $N=8$ subjects (Figure~\ref{fig:metric_agreement_subject_count}), holistic preference reward models perform near random guessing (48.6\% to 57.4\% pairwise agreement), while several reference-similarity metrics lose agreement as subject count increases.

Addressing this gap requires a benchmark and evaluator tailored to multi-subject binding, yet existing benchmarks are under-equipped, focusing on single-subject or limited multi-concept fidelity~\citep{ruiz2023dreambooth,kumari2023multi,peng2025dreambench}, text-only composition~\citep{huang2023t2i}, attribute-level misbinding~\citep{tian2026multibind}, or broad multi-subject capability~\citep{chen2025xverse,wang2026psr,wu2025omnigen2,oshima2026multibanana,borse2025multihuman,xu2026withanyone}, without factorizing the interaction and occlusion conditions that induce binding failures (Appendix~\ref{sec:appendix_related_work}, Table~\ref{tab:benchmark_comparison}). Concretely, existing benchmarks largely lack human pairwise preference labels for reliable cross-generator ranking. Binding is therefore both a generation and a measurement bottleneck.

\begin{figure*}[!ht]
    \centering
    \includegraphics[
        width=0.94\textwidth
    ]{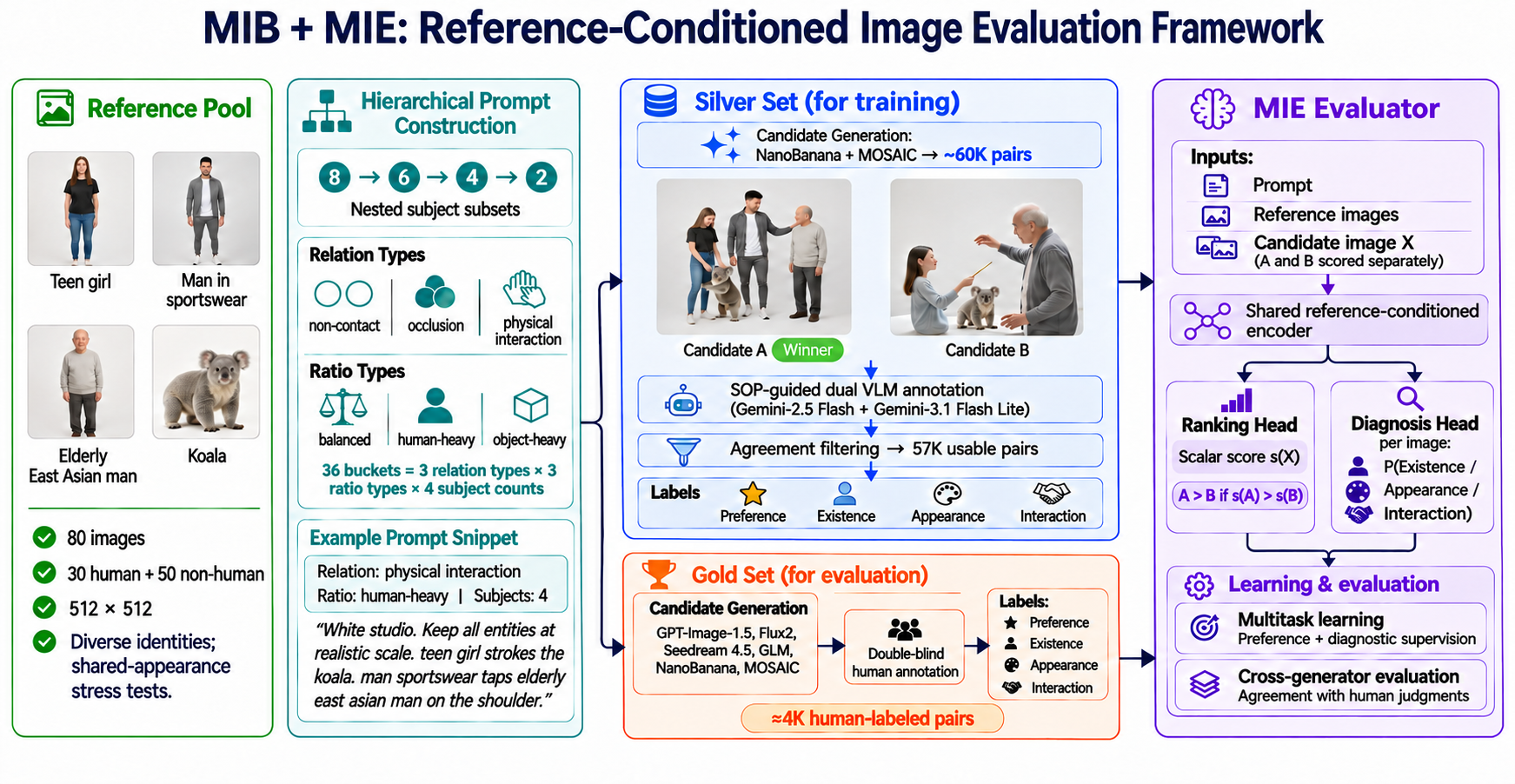}
    \caption{Overview of MIBE. MIB constructs a controlled benchmark through reference pooling, hierarchical prompt construction, Silver Set supervision, and Gold Set human evaluation. MIE is trained on Silver labels and evaluated against double-blind human Gold labels for cross-generator, human-aligned assessment.}
    \label{fig:mibe_overview}
\end{figure*}

To bridge both the benchmarking and evaluation gaps, we introduce MIBE, a unified framework comprising the Multi-subject Interaction Benchmark (MIB) and the Multi-subject Interaction Evaluator (MIE) (Figure~\ref{fig:mibe_overview}). MIB is constructed around a decoupled experimental design that separates reference identity curation, hierarchical prompt factorization, candidate pair generation, and annotation acquisition. To support scalable evaluator learning alongside human evaluation, MIB pairs MIB-Silver, consisting of 60{,}000 candidate pairs (56{,}909 usable consensus pairs) labeled by two frontier VLM judges following a standard operating procedure (SOP), with MIB-Gold, a 4{,}020-pair benchmark (3{,}690 pairs with a consensus winner) annotated by human judges under a double-blind protocol across six generators, four of which are absent from MIB-Silver. Using MIB-Silver, we train MIE, a lightweight, reference-conditioned evaluator that maps the reference set, prompt, and candidate image to both a continuous scalar ranking score and structured diagnostic probabilities over Existence, Appearance, and Interaction. Without any target-generator fine-tuning, MIE achieves 0.922 overall pairwise accuracy against human consensus on MIB-Gold (0.982 on seen generators and 0.884 on unseen generators) and a diagnostic macro-F1 of 0.818. Within each generator matchup, its score margins separate human winners from losers (ROC-AUC 0.77--0.89), and a separate study with real-world photographic references gives preliminary evidence of transfer beyond the reference pool (Appendix~\ref{appendix:robust}).

In summary, our primary contributions are threefold:

 1. We construct MIB, a controlled benchmark for reference-conditioned multi-subject generation that factorizes subject count ($N \in \{2, 4, 6, 8\}$), human/non-human composition, and relation type, pairing a 60K-pair SOP-guided dual-VLM Silver Set with a 4{,}020-pair double-blind human-annotated Gold Set across six generators.

 2. We develop MIE, a parameter-efficient dual-head evaluator trained exclusively on MIB-Silver that jointly predicts global pairwise preference rankings and interpretable facet-level diagnostics across \emph{Existence}, \emph{Appearance}, and \emph{Interaction}, and evaluate its transfer to unseen generators and to novel real-photo references.
 
  3. Through evaluation on MIB-Gold, we show that existing metrics align weakly with human judgment as relational complexity grows, establish MIE's alignment with human preference and diagnostic judgments, and characterize key generative bottlenecks, including coupled Existence and Appearance degradation, figure splitting, and action dropout.

\section{Related Work}

\textbf{Personalized and Multi-Subject Image Generation.}
Personalized text-to-image synthesis has evolved from test-time optimization methods such as Textual Inversion~\citep{gal2022image}, DreamBooth~\citep{ruiz2023dreambooth}, and Custom Diffusion~\citep{kumari2023multi} to feed-forward encoder-based frameworks including IP-Adapter~\citep{ye2023ip}, InstantID~\citep{wang2024instantid}, and PhotoMaker~\citep{li2024photomaker}. While these approaches achieve high identity fidelity for single subjects, extending conditioning to multiple distinct references introduces severe cross-subject interference. Recent multi-subject architectures, including FastComposer~\citep{xiao2025fastcomposer}, Mix-of-Show~\citep{gu2023mix}, MS-Diffusion~\citep{wang2025ms}, UniPortrait~\citep{he2025uniportrait}, AnyStory~\citep{he2025anystory}, MOSAIC~\citep{she2026mosaic}, XVerse~\citep{chen2025xverse}, FocusDPO~\citep{jin2026focusdpo}, PSR~\citep{wang2026psr}, and Interact-Custom~\citep{xu2025interact}, mitigate identity blending through localized cross-attention, regional routing, correspondence-aware disentanglement, or preference optimization. Nevertheless, existing generators are rarely stress-tested in contact-rich or occlusion-heavy regimes where multiple entities overlap spatially, leaving failure modes such as identity fusion and relational misattribution poorly quantified.

\textbf{Benchmarks for Compositional and Personalized Generation.}
Compositional text-to-image benchmarks such as T2I-CompBench~\citep{huang2023t2i} evaluate attribute and spatial alignment from text alone, lacking the reference conditioning required to assess identity preservation. Conversely, early personalization benchmarks including DreamBench~\citep{ruiz2023dreambooth}, CustomConcept101~\citep{kumari2023multi}, and DreamBench++~\citep{peng2025dreambench} focus primarily on single-subject or two-concept scenes. Recent multi-reference benchmarks, such as XVerseBench~\citep{chen2025xverse}, PSRBench~\citep{wang2026psr}, OmniContext~\citep{wu2025omnigen2}, MultiBanana~\citep{oshima2026multibanana}, MultiHuman-Testbench~\citep{borse2025multihuman}, and WithAnyone~\citep{xu2026withanyone}, expand evaluation to multiple subjects, while MultiBind~\citep{tian2026multibind} diagnoses slot-level attribute misbinding. Chen et al.~\citep{chen2026identities} already vary subject count and relation regime. MIB adds nested prompts, composition buckets, scalable VLM supervision, and double-blind human pairwise evaluation across seen and unseen generators (Appendix~\ref{sec:appendix_related_work}, Table~\ref{tab:benchmark_comparison}).

\textbf{Evaluation Metrics and Learned Multimodal Judges.}
Prevailing evaluation protocols rely on feature similarity in pretrained embedding spaces (e.g., CLIP~\citep{radford2021learning}, SigLIP~\citep{zhai2023sigmoid}, DINOv2~\citep{oquab2023dinov2}), bipartite face-identity matching~\citep{deng2019arcface,borse2025multihuman}, or general text-to-image reward models trained on holistic human preferences (e.g., PickScore~\citep{kirstain2023pick}, ImageReward~\citep{xu2023imagereward}, HPS v2.1~\citep{wu2023human}). As we show empirically, general-purpose similarity and preference metrics align weakly with human judgments of binding. Off-the-shelf Vision-Language Models (VLMs) are increasingly deployed as automated judges; they can preserve ranking correlation while assigning unreliable absolute scores~\citep{kumar2026vlm}. MIBE addresses these limitations by formulating supervision around pairwise preference comparisons paired with a structured diagnostic rubric, distilling dual-VLM consensus into a compact, reference-conditioned evaluator validated against double-blind human consensus.

\section{Methods}
\label{sec:method}

\subsection{MIB Construction}

MIB serves as the data foundation of MIBE, comprising controlled prompt-reference tasks, paired candidate generations, and multi-dimensional preference and diagnostic annotations. Each benchmark instance specifies a guiding prompt $p$, a set of $N$ reference subject images $R = \{r_1, \dots, r_N\}$, and two candidate images $(y_A, y_B)$ synthesized under the identical conditioning pair $(R, p)$. As illustrated in Figure~\ref{fig:mibe_overview}, MIB is organized around a decoupled four-stage pipeline that separates reference pool curation, hierarchical prompt factorization, candidate generation, and label acquisition.

\subsubsection{Prompt Construction}
\label{Prompt}

To isolate the effect of increasing relational load from topic or vocabulary shifts, we construct 60{,}000 prompt-reference tasks from 15{,}000 hierarchical Level-8 seeds (Figure~\ref{fig:hierarchical_prompt_construction}). Each seed begins as an eight-entity scene with four core relational clauses and is deterministically reduced into strict Level-6, Level-4, and Level-2 subsets by removing pairs of entities along with their associated relations (see representative examples in Appendix~\ref{sec:app_example_grid}, Table~\ref{tab:structured_example_grid}). Because lower-complexity prompts remain exact semantic subsets of higher-complexity scenes, performance differences across levels reflect the joint increase in subjects and relational constraints rather than changes of topic or vocabulary.

We factorize the prompt space into $4 \times 3 \times 3 = 36$ balanced buckets defined by three orthogonal axes:
(1)~\emph{Subject Count} $N \in \{2, 4, 6, 8\}$;
(2)~\emph{Entity Composition Ratio}, spanning \emph{Balanced}, \emph{Human-heavy}, and \emph{Object-heavy} combinations of human ($\text{H}$) and non-human ($\text{N}$) entities according to the templates in Figure~\ref{fig:hierarchical_prompt_construction}; and
(3)~\emph{Relation Type}, categorized into three progressively demanding interaction regimes: \emph{Non-contact relations,} spatial proximity or perceptual orientation without physical overlap (e.g., standing beside, facing, or glancing at another entity); \emph{Occlusion relations,} partial visibility constraints in which one entity stands behind or partially blocks another without active physical manipulation, testing instance separation under ambiguous visual boundaries; \emph{Physical interactions,} direct subject-to-subject contact (e.g., hugging, shaking hands, petting) or object-mediated actions (e.g., holding, carrying, feeding), requiring precise geometric and role binding.

When multiple relation types co-occur within a multi-entity prompt, the task is assigned to the most restrictive category under the priority order $\text{physical interaction} > \text{occlusion} > \text{non-contact}$. To keep role attribution unambiguous, each entity in a hierarchical seed participates in at most one core relation clause; the separately drawn seen-generator Gold prompts do not strictly enforce this rule. Across the 60{,}000 MIB-Silver prompts, tasks are distributed uniformly with 15{,}000 prompts per subject count, 20{,}000 per composition ratio, and 20{,}004 / 20{,}004 / 19{,}992 across non-contact, occlusion, and physical-interaction categories (1{,}666--1{,}667 prompts per bucket). MIB-Gold follows the same stratified design across 1{,}260 core prompts (315 per subject count, balanced across relation and composition types), supplemented by 1{,}500 stratified tasks for seen-generator evaluation.

To prevent background clutter, lighting variations, or implausible object proportions from confounding binding evaluation, all prompts specify a white studio (\textit{``White studio.''}), and the Silver prompts, including those reused for unseen-generator Gold, also require realistic entity scale (\textit{``Keep all entities at realistic scale.''}). Each hierarchical Silver prompt undergoes rule-based checks to enforce seven structural invariants: (i)~all assigned entities are explicitly mentioned, (ii)~unassigned entities are excluded, (iii)~human and non-human counts match the target composition bucket, (iv)~lower-complexity prompts form strict subsets of their parent seed, (v)~duplicate prompts are removed, (vi)~no entity serves as the grammatical subject of multiple independent relations, and (vii)~physical-contact verbs are excluded from non-contact buckets.

\subsubsection{Reference and Candidate Image Generation}

MIB utilizes a curated pool of 80 synthetic reference subjects standardized as isolated $512 \times 512$ studio images generated via GPT-Image-1, comprising 30 human identities and 50 non-human categories (animals, foods, furniture, tools, wearables, and manipulable objects; full demographic and categorical breakdown in Appendix~\ref{appendix:reference_pool}, Table~\ref{tab:reference_distribution}). Human references span diverse age bands, gender presentations, ethnicities, hairstyles, and garments. To prevent generators and evaluators from relying on coarse attire shortcuts rather than fine-grained facial and structural identity, half of the human subjects share visually similar casual wear (e.g., gray T-shirts and jeans) and co-occur more often as subject count rises from $N=2$ to $N=8$ (Appendix~\ref{sec:appendix_failure_modes}, Table~\ref{tab:similarity_cooccurrence}).

For each prompt-reference task, two personalized generators receive the identical guiding prompt $p$ and reference image set $R$ to synthesize a candidate pair $(y_A, y_B)$, ensuring that pairwise comparisons measure relative binding quality under identical conditioning. To decouple scalable evaluator training from cross-generator benchmarking, we construct two complementary dataset splits (supplementary Table~\ref{tab:dataset_statistics}):

\emph{MIB-Silver (Training Set):} We generate 60{,}000 candidate pairs using Nano Banana~\citep{deepmind2025nanobanana} and MOSAIC~\citep{she2026mosaic}. We selected this generator pairing because preliminary inspection demonstrated strongly complementary failure profiles, including subject omission, identity drift, cross-subject appearance bleeding, and relation hallucination, yielding informative pairwise contrast across Existence, Appearance, and Interaction. Following consensus filtering (Section~\ref{sec:method_label}), the usable Silver tasks are partitioned at the Level-8 seed-family level into disjoint training, validation, and test splits so that no seed family spans multiple splits.

\emph{MIB-Gold (Human Evaluation Set):} To test whether evaluation metrics generalize to unseen generator distributions rather than overfitting to training-generator artifacts, MIB-Gold comprises 4{,}020 human-annotated candidate pairs across six generators divided into two evaluation subsets: (i)~an \emph{unseen-generator split} of 2{,}520 pairs, constructed by rendering 1{,}260 stratified prompts across four held-out generators that never appear during MIE training (GPT-Image-1.5~\citep{openai2025gptimage15}, Flux.2~\citep{flux-2-2025}, Seedream~4.5~\citep{seed2026seedream}, and GLM~\citep{zai2026glmimage}), organized into two fixed cross-model matchups (GPT-Image-1.5 vs.\ Flux.2 and Seedream~4.5 vs.\ GLM); these prompts are re-rendered from MIB-Silver and share its reference pool, so this split tests unseen generators rather than unseen prompts or identities (overlap protocol in supplementary Sec.~\ref{sec:app_overlap}); and (ii)~a \emph{seen-generator split} of 1{,}500 pairs generated by Nano Banana and MOSAIC on a separately drawn set of stratified prompts. Prior to annotation, candidate order ($A$ vs.\ $B$) is randomized and corrupted or blank generations are removed via automated validity checks.

\subsubsection{Label Construction}
\label{sec:method_label}

MIB provides two complementary annotation signals for each candidate pair: a binary \emph{pairwise preference label} for global ranking and three-dimensional \emph{diagnostic labels} for failure localization. Pairwise preference asks which candidate ($y_A$ or $y_B$) is superior overall under the shared condition $(R, p)$, capturing relative failure severity when both candidates exhibit imperfections. To localize \emph{why} a generation succeeds or fails, each candidate image is independently annotated along three diagnostic dimensions: 1) \emph{Existence:} Whether all requested reference subjects are physically present as distinct entities without omission, cloning, or unidentifiable collapse. 2) \emph{Appearance:} Whether rendered subjects preserve the structural geometry, facial identity, and wardrobe attributes of their references without severe distortion or cross-subject feature bleeding. 3) \emph{Interaction:} Whether subject-to-subject spatial relations, occlusion ordering, actions, and object assignments match the prompt and are physically plausible.

Because global ranking requires an unambiguous target, pairs whose two judgments name different winners are discarded. For per-image diagnostic dimensions, where borderline visual defects naturally induce graded uncertainty, we average the two binary annotations into soft supervision scores in $\{0, 0.5, 1\}$.

\textbf{MIB-Silver Annotation via SOP-Guided Dual-VLM Consensus.}
To obtain scalable supervision without collapsing into holistic aesthetic judgments, we enforce a unified Standard Operating Procedure (SOP; full prompt in Appendix~\ref{sec:app_SOP}) across two independent VLM judges: gemini-2.5-flash~\citep{comanici2025gemini} and gemini-3.1-flash-lite-preview. The SOP enforces a structured reasoning chain: before rendering a pairwise preference, each judge must independently inspect candidates $y_A$ and $y_B$, generate explicit subject-level flaw logs documenting missing entities, wardrobe/identity mismatches, or spatial/physical violations, and assign binary scores across Existence, Appearance, and Interaction (qualitative examples in Appendix~\ref{sec:app_SOP_demo}, Figure~\ref{fig:sop_demos}). Across 59{,}852 valid Silver comparisons, the two VLM judges agree on the preferred candidate in 95.1\% of cases (Cohen's $\kappa = 0.37$, low because both judges prefer the same generator in most tasks), and per-dimension agreement on both images of a pair is 89.1\% on Existence, 71.9\% on Interaction, and 69.0\% on Appearance. Consensus filtering yields 56{,}909 Silver pairs whose per-dimension diagnostic scores are averaged as soft training targets; the released splits additionally contain 146 records labeled by a single judge (129 in the training split).

\textbf{MIB-Gold Annotation via Double-Blind Human Evaluation.}
MIB-Gold is annotated exclusively by trained human judges following the identical three-dimensional SOP under a strict double-blind protocol: each of the 4{,}020 candidate pairs is independently evaluated by two annotators blinded to generator identity and each other's ratings, with candidate order ($A$/$B$) randomized. Enforcing strict consensus on the pairwise winner retains 3{,}690 preference-consistent evaluation pairs overall (91.8\% retention), comprising 1{,}411 of 1{,}500 pairs (94.1\%) on the seen-generator split and 2{,}279 of 2{,}520 pairs (90.4\%) on the unseen-generator split, while preserving substantial diagnostic challenge across Existence, Appearance, and Interaction. Per-dimension inter-annotator agreement on the 8{,}040 Gold images is Fleiss' $\kappa = 0.85$ / $0.56$ / $0.47$ for Existence / Appearance / Interaction.

\subsection{Evaluator Modeling}

To make MIB usable for automatic evaluation, we develop the Multi-subject Interaction Evaluator (MIE), a lightweight, reference-conditioned multimodal evaluator trained exclusively on MIB-Silver and evaluated zero-shot on MIB-Gold.

\textbf{Problem Formulation.}
Given a guiding prompt $p$, a set of $N$ reference subject images $R = \{r_1, \dots, r_N\}$, and a candidate generated image $y$, MIE parameterizes a mapping $(R, p, y) \mapsto (s_\theta, \hat{d}_\theta)$ producing two complementary outputs:
(1)~a scalar utility score $s_\theta(R, p, y) \in \mathbb{R}$ used for continuous pairwise preference ranking, and
(2)~a three-dimensional diagnostic logit vector $\hat{d}_\theta(R, p, y) = (\hat{d}_{\text{exist}}, \hat{d}_{\text{app}}, \hat{d}_{\text{inter}}) \in \mathbb{R}^3$, whose element-wise sigmoid probabilities $\sigma(\hat{d}_\theta(R, p, y)) \in [0, 1]^3$ estimate whether $y$ satisfies \emph{Existence}, \emph{Appearance}, and \emph{Interaction}.

\textbf{Multimodal Input and Scoring.}
Rather than relying on autoregressive text generation or brittle in-context scoring rubrics at inference time, MIE operates as a direct discriminative read-out model atop a pretrained vision-language backbone (Qwen3.5-VLM~\citep{qwen2026qwen35}). For a task with $N$ reference subjects, the input is formatted as a single-turn multimodal sequence containing $N$ reference-image placeholders, one candidate-image placeholder, and a fixed instruction template: \textit{``You are evaluating a multi-subject personalization result. The first images are subject references. The last image is the generated candidate. Prompt: \{prompt\}''}, processed with \texttt{add\_generation\_prompt=False}.  When scoring a candidate pair $(y_A, y_B)$ conditioned on shared $(R, p)$, the shared references and prompt are loaded and serialized once, and $y_A$ and $y_B$ are each scored in a single forward pass. From the final transformer layer, we extract the last non-padding token's hidden state $h_y$ for each candidate $y \in \{y_A, y_B\}$ and project it through two lightweight MLP prediction heads to obtain $s_\theta(R, p, y)$ and $\hat{d}_\theta(R, p, y)$. The predicted pairwise winner is determined by the score margin $\hat{y}_{(A \succ B)} = \mathrm{sign}(s_\theta(R, p, y_A) - s_\theta(R, p, y_B))$.

\textbf{Dual-Head Architecture and Multi-Task Objective.}
Relying solely on binary diagnostic classifiers cannot resolve relative quality when two candidates fail the same dimension to different degrees (e.g., omitting one subject versus omitting four), whereas a standalone scalar reward model provides an opaque score without localizing the source of failure. MIE bridges both requirements through its dual-head architecture: the \emph{ranking head} produces a continuous global utility for pairwise leaderboard comparison, while the \emph{diagnostic head} attributes errors across Existence, Appearance, and Interaction. We optimize both heads jointly via a multi-task objective:
\begin{equation}
    \mathcal{L} = \alpha \mathcal{L}_{\text{rank}} + \beta \mathcal{L}_{\text{diag}},
\end{equation}
where $\mathcal{L}_{\text{rank}}$ is a margin ranking loss ($\text{margin} = 0.1$, weight $\alpha = 1.0$) supervised by the consensus pairwise preference label, and $\mathcal{L}_{\text{diag}}$ is a binary cross-entropy loss with logits (weight $\beta = 0.5$) supervised by the soft per-dimension diagnostic targets across both candidates (Appendix~\ref{sec:appendix_configuration}, Table~\ref{tab:loss_config}). This joint formulation mirrors our human annotation protocol, training a single shared representation to support both pairwise preference discrimination and facet-level diagnosis.

\textbf{Parameter-Efficient Adaptation.}
To ensure that MIE remains computationally accessible and to study how benchmark supervision interacts with model capacity and adaptation strategy, we fine-tune off-the-shelf Qwen3.5-VLM backbones across three parameter scales (0.8B, 2B, and 4B). For each scale, we compare unfreezing only the top four backbone layers (\emph{layer-only}) against combining top-layer unfreezing with low-rank adapters~\citep{hu2022lora} (\emph{LoRA-layer}, $r=16, \alpha=32$ on attention and MLP projections; Table~\ref{tab:training_hparams}) under a standardized training budget of 600 optimization steps (9{,}600 Silver pairs). As we show in Section~4.2, adapting a compact 4B backbone on this 9.6K-pair subset of MIB-Silver is sufficient to achieve strong alignment with human preference and diagnostic labels across both seen and unseen generators.

\section{Results}
Training, loss, and evaluation configurations are detailed in Appendix~\ref{sec:appendix_configuration}.

\begin{table*}[t]
\centering
\small
\caption{Human alignment of MIE (4B, LoRA-layer) on the 3{,}690 consensus pairs of MIB-Gold. CIs: 95\% bootstrap over prompt-seed groups (4{,}000 resamples). Winner share: fraction of pairs won by the matchup's dominant generator. Minority: correctly ranked pairs among those won by the other generator. Held-out seeds: unseen-split pairs whose prompt seed group was never in MIE's training split.}
\label{tab:validation}
\begin{tabular}{@{}lrcccc@{}}
\toprule
Gold pairs & $n$ & Acc. (95\% CI) & Winner share & AUC & Minority \\
\midrule
Nano Banana vs.\ MOSAIC (seen) & 1,411 & 0.982 (0.974--0.989) & 98.8\% & 0.864 & 4/17 \\
GPT-Image-1.5 vs.\ Flux.2 (unseen) & 1,071 & 0.779 (0.752--0.803) & 90.5\% & 0.766 & 54/102 \\
Seedream 4.5 vs.\ GLM (unseen) & 1,208 & 0.978 (0.969--0.986) & 98.3\% & 0.892 & 4/20 \\
\midrule
All unseen & 2,279 & 0.884 (0.871--0.897) & -- & -- & 58/122 \\
\quad held-out seeds & 286 & 0.895 (0.861--0.928) & -- & -- & -- \\
Overall & 3,690 & 0.922 (0.913--0.931) & -- & -- & 62/139 \\
\bottomrule
\end{tabular}
\end{table*}
\subsection{Existing Metrics Align Weakly with Humans}

We first investigate whether existing automatic evaluation metrics can serve as reliable proxies for human judgment on MIB-Gold. We evaluate 12 standard baselines spanning four families:
(1)~\emph{Low-level reconstruction metrics:} PSNR~\citep{hore2010image}, SSIM~\citep{wang2004image}, and MS-SSIM~\citep{wang2003multiscale}, which quantify pixel- and structure-level similarity;
(2)~\emph{Semantic and reference alignment metrics:} DINOv2~\citep{oquab2023dinov2}, CLIP~\citep{radford2021learning}, and SigLIP~\citep{zhai2023sigmoid} (CLIP/SigLIP in text-image (-T) and reference-image (-I) modes) to measure semantic consistency and global identity similarity;
(3)~\emph{General human preference models:} PickScore~\citep{kirstain2023pick}, ImageReward~\citep{xu2023imagereward}, and HPS v2.1~\citep{wu2023human}, which are trained on human feedback for general text-to-image aesthetics and prompt alignment; and
(4)~\emph{Identity-specific metrics:} Subject Collapse Rate (SCR)~\citep{chen2026identities}, designed to assess identity preservation under multi-subject stress.

As shown in Figure~\ref{fig:metric_agreement_subject_count}, evaluating these baselines against double-blind human preferences (50\% is chance) shows weak agreement across all four metric families. On the unseen split, PickScore and HPS v2.1 reach 48.6\% and 52.0\%, and the reconstruction metric PSNR only 39.9\%. In Figure~\ref{fig:metric_agreement_subject_count}(b), SigLIP-I and SSIM agree less with humans in eight-subject scenes than in two-subject scenes. Scored on the same 4{,}991 individual unseen-split annotator judgments (including pairs without a consensus winner), with ties counted as half agreement for every method, MIE agrees with humans on 84.7\%, 21.0 points above the best of these twelve metrics (DINO, 63.6\%); on the seen-generator split the figures are 95.4\% and 74.6\% (MS-SSIM). Face-identity matching is a strong signal where it applies: on the 4{,}573 unseen-split judgments that involve human references, ArcFace distance~\citep{deng2019arcface} reaches 87.2\%, against 85.7\% for MIE; ArcFace's scope is human identity matching, with no non-human or interaction diagnostics.

\begin{table*}[t]
\centering
\small
\caption{Benchmarking six generators on the 7,806 valid Gold Set images. Existence, Appearance, and Interaction report the mean MIE diagnostic probability per dimension; All Satisfied (MIE) is the fraction of images with all three probabilities $>0.5$. Human columns compute All Satisfied from the Gold labels: \emph{both} requires both annotators to accept all three dimensions; \emph{lenient} requires at least one annotator per dimension. The two human columns are aggregation rules, not uncertainty bounds.}
\label{tab:generator_benchmark}
\begin{tabular}{lccccccc}
\toprule
 & & \multicolumn{3}{c}{MIE mean probability} & \multicolumn{3}{c}{All Satisfied} \\
\cmidrule(lr){3-5}\cmidrule(l){6-8}
Generator & $n$ & Exist. & App. & Inter. & MIE & Human (both) & Human (lenient) \\
\midrule
Seedream 4.5      & 1,226 & \textbf{0.925} & 0.674 & 0.477 & 47.2\% & 23.2\% & 49.2\% \\
GPT-Image-1.5      & 1,226 & 0.846 & \textbf{0.684} & 0.491 & \textbf{52.4\%} & \textbf{28.1\%} & \textbf{61.2\%} \\
Nano Banana        & 1,451 & 0.811 & 0.549 & \textbf{0.558} & 42.0\% & 19.0\% & 34.9\% \\
Flux.2-klein-9B    & 1,226 & 0.772 & 0.402 & 0.310 & 22.3\% & 4.6\% & 18.4\% \\
GLM                & 1,226 & 0.402 & 0.148 & 0.079 & 1.6\% & 0.8\% & 4.6\% \\
MOSAIC             & 1,451 & 0.281 & 0.121 & 0.065 & 0.8\% & 0.6\% & 2.6\% \\
\bottomrule
\end{tabular}
\label{table2}
\end{table*}
\subsection{MIE Aligns Better with Human Preference}

Given the weak agreement of existing metrics, we next evaluate whether supervision from MIB-Silver enables MIE to align with human judgments. Across our six trained checkpoints (spanning 0.8B, 2B, and 4B backbones under layer-only and LoRA-layer adaptation), our strongest variant (Qwen3.5-4B with LoRA-layer tuning) achieves an overall pairwise agreement of 0.922 with human consensus on MIB-Gold, reaching 0.982 on the seen-generator split (Nano Banana vs.\ MOSAIC) and 0.884 on the unseen-generator split (GPT-Image-1.5, Flux.2, Seedream~4.5, and GLM). Because one generator wins most pairs within each fixed matchup, accuracy alone overstates instance-level discrimination, so Table~\ref{tab:validation} also reports winner prevalence, ROC-AUC of the score margin, and recall of minority wins.

\textbf{Validation beyond accuracy.} Within every matchup, MIE's score margin separates human winners from losers (ROC-AUC 0.864, 0.766, and 0.892), and MIE recovers 62 of the 139 pairs in which the less-preferred generator wins, including 54 of 102 in the hardest matchup (GPT-Image-1.5 vs.\ Flux.2), where human preferences are least one-sided and annotators disagree most (85.0\% retention). A post hoc rule that always picks the dominant generator matches or exceeds MIE's accuracy in each matchup, but it has AUC 0.5 and recovers none of these pairs. The seen-to-unseen drop is concentrated in this matchup (accuracy 0.779), whereas Seedream~4.5 vs.\ GLM (0.978) is close to the seen matchup (0.982). Because the unseen split reuses Silver prompts, we also score prompts whose seed group was held out of MIE's training split: accuracy is 0.895 ($n{=}286$), against 0.884 on prompts from the training split ($n{=}1{,}883$).

MIE's alignment extends beyond global pairwise ranking to fine-grained failure localization. On the 3{,}690 consensus Gold pairs, using images with agreeing labels per dimension and threshold 0.5 (supplementary Sec.~\ref{sec:app_diag}), the 4B LoRA-layer model achieves a macro-F1 of 0.818 (F1 0.927 / 0.784 / 0.744 for Existence / Appearance / Interaction). As shown in Figure~\ref{fig:mie_alignment} (supplementary material), checkpoints with higher pairwise accuracy generally also achieve higher per-category F1. Table~\ref{tab:lora_ablation} in the supplementary material summarizes the comparison across all six model variants. Across model scales, LoRA-layer adaptation improves unseen-split accuracy over layer-only tuning for the 2B and 4B backbones and improves macro-F1 at every scale, so we use it by default (Table~\ref{tab:training_hparams}).

\textbf{Robustness to Unseen References.}
Appendix~\ref{appendix:robust} tests the unchanged MIE checkpoint on 22 unseen subjects: 10 human identities and 12 non-human categories. On GPT-Image-1.5 versus Flux.2, AUC is 0.827 with real-photo references and 0.851 and 0.775 with GPT-Image-1 and Qwen-Image references (Table~\ref{tab:unseen_class_auc}), providing preliminary evidence of ranking transfer; Existence correlations vary by generator. On a per-subject difficulty index, the tested novel non-human categories score higher (easier) than the novel human identities (Table~\ref{tab:longtail_difficulty}); this comparison does not isolate category novelty.

\textbf{Benchmarking Generators on MIB.}
Finally, we apply MIE to benchmark all six generative models across the 7{,}806 valid candidate images in MIB-Gold (Table~\ref{table2}). For each image, MIE outputs facet-level satisfaction probabilities in $[0, 1]$ across Existence, Appearance, and Interaction; because a multi-subject synthesis is fully successful only when all three binding criteria hold simultaneously, we also report the \emph{All Satisfied} rate (the proportion of images achieving $> 0.5$ across all three dimensions). Because MIE's probabilities are not calibrated success rates, Table~\ref{table2} also reports All Satisfied computed directly from the two Gold annotators' diagnostic labels on the same images. MIE ranks the six generators exactly as both human aggregation rules do (Spearman $\rho = 1.0$), and across generators its per-dimension means correlate with the human means at Pearson $r = 0.95$ / $0.96$ / $0.96$. Even the strongest generator, GPT-Image-1.5, satisfies all three criteria for both annotators in only 28.1\% of images (52.4\% according to MIE).

Human labels show where generators fail (Appendix~\ref{sec:appendix_failure_modes}). \emph{Existence} is the most often satisfied dimension (human mean up to 0.865 for Seedream~4.5). \emph{Appearance} is the weakest dimension for four of the six generators (Nano Banana, Flux.2, GLM, and MOSAIC; human means 0.07--0.40): when a generator omits a subject, it can blend that subject's garments or facial traits onto surviving figures (Figure~\ref{fig:appendix-existence-appearance-cooccurrence}). \emph{Interaction} is the weakest dimension for GPT-Image-1.5 and Seedream~4.5. Figure~\ref{fig:appendix-multi-action-overload} illustrates \emph{figure splitting}, duplicating an identity to satisfy concurrent actions, and \emph{action dropout}, silently discarding assigned relations, in Nano Banana. MIE reproduces the generator ordering within each dimension closely (rank correlations 0.94 / 0.94 / 0.89) but tends to overestimate Existence and Appearance, so the per-dimension conclusions above rest on the human labels.

\textbf{Limitations and Broader Impacts.}
Several scope constraints are discussed further in Appendix~\ref{appendix:limit}. First, MIB uses a closed pool of 80 synthetic studio references to isolate relational binding from background confounders; although Appendix~\ref{appendix:robust} provides preliminary evidence of transfer to 22 real-world photographic subjects, extending to unconstrained scenes remains important future work. Second, Silver uses two generators and two VLM judges; teacher--human diagnostic agreement and sensitivity to alternative teachers are unmeasured. Third, single-reference conditioning cannot fully separate pose-invariant identity preservation from direct visual replication~\citep{xu2026withanyone}, motivating multi-reference extensions. Fourth, each Gold matchup is fixed and one generator wins most of its pairs, so accuracy should be read together with AUC and minority-win recall (Table~\ref{tab:validation}); each MIE checkpoint was trained once (seed 3407), and the contribution of the diagnostic loss to ranking is not isolated.

\section{Conclusion}
We presented MIBE, a benchmark and evaluator for subject binding in multi-subject personalized generation. MIB varies subject count, human/non-human composition, and relation type, and pairs a 60K-pair dual-VLM Silver Set with 4{,}020 double-blind human-labeled Gold pairs from six generators. Trained only on Silver, MIE agrees with human consensus on 92.2\% of consensus Gold pairs (88.4\% on unseen generators), separates human winners from losers within every matchup (ROC-AUC 0.77--0.89), and ranks the six generators by All Satisfied rate exactly as human labels do. Data, annotations, and code will be released.

\section*{LLM Usage Statement}
Generative AI is part of the research methodology of this work. (i)~\emph{Synthetic data}: all 80 benchmark reference images were generated with GPT-Image-1, and all candidate images were produced by the generators under study (Nano Banana, MOSAIC, GPT-Image-1.5, Flux.2, Seedream~4.5, and GLM). (ii)~\emph{Labeling}: MIB-Silver preference and diagnostic labels were produced by gemini-2.5-flash and gemini-3.1-flash-lite-preview following the SOP in the supplementary material, and the combinatorial prompt-generation and rule-based validation scripts were written and run with AI coding assistants (Claude Code and OpenAI Codex). All MIB-Gold labels used for evaluation were produced by human annotators. (iii)~\emph{Evaluator}: MIE fine-tunes open Qwen3.5 vision-language backbones on MIB-Silver. (iv)~\emph{Code and writing}: AI coding assistants (Anthropic Claude via Claude Code, and OpenAI Codex) helped write and edit experiment, analysis, and figure-generation code; AI assistants (Claude, and OpenAI Codex for pre-submission review) were used to edit the manuscript for readability, check numbers against result files, and format the bibliography. All AI-assisted code was reviewed and run by the authors, and every bibliography entry was checked by the authors against its original source. The authors take full responsibility for all content, including text, claims, and artifacts produced with generative AI.

{
    \small
    \bibliographystyle{ieeenat_fullname}
    \bibliography{ref}

\begin{thebibliography}{43}
\providecommand{\natexlab}[1]{#1}
\providecommand{\url}[1]{\texttt{#1}}
\expandafter\ifx\csname urlstyle\endcsname\relax
  \providecommand{\doi}[1]{doi: #1}\else
  \providecommand{\doi}{doi: \begingroup \urlstyle{rm}\Url}\fi

\bibitem[{Black Forest Labs}(2025)]{flux-2-2025}
{Black Forest Labs}.
\newblock {FLUX.2: Frontier Visual Intelligence}.
\newblock \url{https://bfl.ai/blog/flux-2}, 2025.

\bibitem[Borse et~al.(2025)Borse, Choi, Park, Kim, Kadambi, Garrepalli, Yun, Malladi, and Porikli]{borse2025multihuman}
Shubhankar Borse, Seokeon Choi, Sunghyun Park, Jeongho Kim, Shreya Kadambi, Risheek Garrepalli, Sungrack Yun, Durga Malladi, and Fatih Porikli.
\newblock Multihuman-testbench: Benchmarking image generation for multiple humans.
\newblock In \emph{Advances in Neural Information Processing Systems}, 2025.

\bibitem[{ByteDance Seed Team}(2025)]{seed2026seedream}
{ByteDance Seed Team}.
\newblock {Seedream 4.5}.
\newblock \url{https://seed.bytedance.com/en/seedream4_5}, 2025.

\bibitem[Chen et~al.(2025)Chen, Zhao, Sun, Chen, Wang, Du, and Wu]{chen2025xverse}
Bowen Chen, Mengyi Zhao, Haomiao Sun, Li Chen, Xu Wang, Kang Du, and Xinglong Wu.
\newblock Xverse: Consistent multi-subject control of identity and semantic attributes via dit modulation.
\newblock In \emph{Advances in Neural Information Processing Systems}, pages 7437--7459, 2025.

\bibitem[Chen et~al.(2026)Chen, Zhao, Zhu, and Yao]{chen2026identities}
Zhihan Chen, Yuhuan Zhao, Yijie Zhu, and Xinyu Yao.
\newblock When identities collapse: A stress-test benchmark for multi-subject personalization.
\newblock \emph{arXiv preprint arXiv:2603.26078}, 2026.

\bibitem[Comanici et~al.(2025)Comanici, Bieber, Schaekermann, Pasupat, Sachdeva, Dhillon, Blistein, Ram, Zhang, Rosen, et~al.]{comanici2025gemini}
Gheorghe Comanici, Eric Bieber, Mike Schaekermann, Ice Pasupat, Noveen Sachdeva, Inderjit Dhillon, Marcel Blistein, Ori Ram, Dan Zhang, Evan Rosen, et~al.
\newblock Gemini 2.5: Pushing the frontier with advanced reasoning, multimodality, long context, and next generation agentic capabilities.
\newblock \emph{arXiv preprint arXiv:2507.06261}, 2025.

\bibitem[Deng et~al.(2019)Deng, Guo, Xue, and Zafeiriou]{deng2019arcface}
Jiankang Deng, Jia Guo, Niannan Xue, and Stefanos Zafeiriou.
\newblock Arcface: Additive angular margin loss for deep face recognition.
\newblock In \emph{Proceedings of the IEEE/CVF conference on computer vision and pattern recognition}, pages 4690--4699, 2019.

\bibitem[Fortin et~al.(2025)Fortin, Vernade, Kampf, and Reshi]{deepmind2025nanobanana}
Alisa Fortin, Guillaume Vernade, Kat Kampf, and Ammaar Reshi.
\newblock Introducing {Gemini 2.5 Flash Image}, our state-of-the-art image model.
\newblock \url{https://developers.googleblog.com/en/introducing-gemini-2-5-flash-image/}, 2025.

\bibitem[Gal et~al.(2022)Gal, Alaluf, Atzmon, Patashnik, Bermano, Chechik, and Cohen-Or]{gal2022image}
Rinon Gal, Yuval Alaluf, Yuval Atzmon, Or Patashnik, Amit~H Bermano, Gal Chechik, and Daniel Cohen-Or.
\newblock An image is worth one word: Personalizing text-to-image generation using textual inversion.
\newblock \emph{arXiv preprint arXiv:2208.01618}, 2022.

\bibitem[Gu et~al.(2023)Gu, Wang, Wu, Shi, Chen, Fan, Xiao, Zhao, Chang, Wu, et~al.]{gu2023mix}
Yuchao Gu, Xintao Wang, Jay~Zhangjie Wu, Yujun Shi, Yunpeng Chen, Zihan Fan, Wuyou Xiao, Rui Zhao, Shuning Chang, Weijia Wu, et~al.
\newblock Mix-of-show: Decentralized low-rank adaptation for multi-concept customization of diffusion models.
\newblock \emph{Advances in neural information processing systems}, 36:\penalty0 15890--15902, 2023.

\bibitem[He et~al.(2025{\natexlab{a}})He, Geng, and Bo]{he2025uniportrait}
Junjie He, Yifeng Geng, and Liefeng Bo.
\newblock Uniportrait: A unified framework for identity-preserving single-and multi-human image personalization.
\newblock In \emph{2025 IEEE/CVF International Conference on Computer Vision (ICCV)}, pages 14399--14408. IEEE, 2025{\natexlab{a}}.

\bibitem[He et~al.(2025{\natexlab{b}})He, Tuo, Chen, Zhong, Geng, and Bo]{he2025anystory}
Junjie He, Yuxiang Tuo, Binghui Chen, Chongyang Zhong, Yifeng Geng, and Liefeng Bo.
\newblock Anystory: Towards unified single and multiple subject personalization in text-to-image generation.
\newblock \emph{arXiv preprint arXiv:2501.09503}, 2025{\natexlab{b}}.

\bibitem[Hore and Ziou(2010)]{hore2010image}
Alain Hore and Djemel Ziou.
\newblock Image quality metrics: Psnr vs. ssim.
\newblock In \emph{2010 20th international conference on pattern recognition}, pages 2366--2369. IEEE, 2010.

\bibitem[Hu et~al.(2022)Hu, Shen, Wallis, Allen-Zhu, Li, Wang, Wang, and Chen]{hu2022lora}
Edward~J Hu, Yelong Shen, Phillip Wallis, Zeyuan Allen-Zhu, Yuanzhi Li, Shean Wang, Lu Wang, and Weizhu Chen.
\newblock {LoRA}: Low-rank adaptation of large language models.
\newblock In \emph{International Conference on Learning Representations}, 2022.

\bibitem[Huang et~al.(2023)Huang, Sun, Xie, Li, and Liu]{huang2023t2i}
Kaiyi Huang, Kaiyue Sun, Enze Xie, Zhenguo Li, and Xihui Liu.
\newblock T2i-compbench: A comprehensive benchmark for open-world compositional text-to-image generation.
\newblock \emph{Advances in Neural Information Processing Systems}, 36:\penalty0 78723--78747, 2023.

\bibitem[Jin et~al.(2026)Jin, Fu, She, Jia, Wang, Liu, and Jiang]{jin2026focusdpo}
Qiaoqiao Jin, Siming Fu, Dong She, Weinan Jia, Hualiang Wang, Mu Liu, and Jidong Jiang.
\newblock Focusdpo: Dynamic preference optimization for multi-subject personalized image generation via adaptive focus.
\newblock In \emph{Proceedings of the AAAI Conference on Artificial Intelligence}, pages 5512--5520, 2026.

\bibitem[Kirstain et~al.(2023)Kirstain, Polyak, Singer, Matiana, Penna, and Levy]{kirstain2023pick}
Yuval Kirstain, Adam Polyak, Uriel Singer, Shahbuland Matiana, Joe Penna, and Omer Levy.
\newblock Pick-a-pic: An open dataset of user preferences for text-to-image generation.
\newblock \emph{Advances in neural information processing systems}, 36:\penalty0 36652--36663, 2023.

\bibitem[Kumar et~al.(2026)Kumar, Tayebati, Naik, Krishnan, and Trivedi]{kumar2026vlm}
Divake Kumar, Sina Tayebati, Devashri Naik, Ranganath Krishnan, and Amit~Ranjan Trivedi.
\newblock Vlm judges can rank but cannot score: Task-dependent uncertainty in multimodal evaluation.
\newblock \emph{arXiv preprint arXiv:2604.25235}, 2026.

\bibitem[Kumari et~al.(2023)Kumari, Zhang, Zhang, Shechtman, and Zhu]{kumari2023multi}
Nupur Kumari, Bingliang Zhang, Richard Zhang, Eli Shechtman, and Jun-Yan Zhu.
\newblock Multi-concept customization of text-to-image diffusion.
\newblock In \emph{2023 IEEE/CVF Conference on Computer Vision and Pattern Recognition (CVPR)}, pages 1931--1941. IEEE, 2023.

\bibitem[Li et~al.(2024)Li, Cao, Wang, Qi, Cheng, and Shan]{li2024photomaker}
Zhen Li, Mingdeng Cao, Xintao Wang, Zhongang Qi, Ming-Ming Cheng, and Ying Shan.
\newblock Photomaker: Customizing realistic human photos via stacked id embedding.
\newblock In \emph{2024 IEEE/CVF Conference on Computer Vision and Pattern Recognition (CVPR)}, pages 8640--8650. IEEE, 2024.

\bibitem[{OpenAI}(2025)]{openai2025gptimage15}
{OpenAI}.
\newblock Gpt-image-1.5 model.
\newblock \url{https://platform.openai.com/docs/models/gpt-image-1.5}, 2025.

\bibitem[Oquab et~al.(2023)Oquab, Darcet, Moutakanni, Vo, Szafraniec, Khalidov, Fernandez, Haziza, Massa, El-Nouby, et~al.]{oquab2023dinov2}
Maxime Oquab, Timoth{\'e}e Darcet, Th{\'e}o Moutakanni, Huy Vo, Marc Szafraniec, Vasil Khalidov, Pierre Fernandez, Daniel Haziza, Francisco Massa, Alaaeldin El-Nouby, et~al.
\newblock Dinov2: Learning robust visual features without supervision.
\newblock \emph{arXiv preprint arXiv:2304.07193}, 2023.

\bibitem[Oshima et~al.(2026)Oshima, Miyake, Matsutani, Iwasawa, Suzuki, Matsuo, and Furuta]{oshima2026multibanana}
Yuta Oshima, Daiki Miyake, Kohsei Matsutani, Yusuke Iwasawa, Masahiro Suzuki, Yutaka Matsuo, and Hiroki Furuta.
\newblock Multibanana: A challenging benchmark for multi-reference text-to-image generation.
\newblock In \emph{Proceedings of the IEEE/CVF Conference on Computer Vision and Pattern Recognition}, pages 448--460, 2026.

\bibitem[Peng et~al.(2025)Peng, Cui, Tang, Qi, Dong, Bai, Han, Ge, Zhang, and Xia]{peng2025dreambench}
Yuang Peng, Yuxin Cui, Haomiao Tang, Zekun Qi, Runpei Dong, Jing Bai, Chunrui Han, Zheng Ge, Xiangyu Zhang, and Shu-Tao Xia.
\newblock Dreambench++: A human-aligned benchmark for personalized image generation.
\newblock In \emph{International Conference on Learning Representations}, pages 46010--46032, 2025.

\bibitem[{Qwen Team}(2026)]{qwen2026qwen35}
{Qwen Team}.
\newblock {Qwen3.5}: Towards native multimodal agents.
\newblock \url{https://qwen.ai/blog?id=qwen3.5}, 2026.

\bibitem[Radford et~al.(2021)Radford, Kim, Hallacy, Ramesh, Goh, Agarwal, Sastry, Askell, Mishkin, Clark, et~al.]{radford2021learning}
Alec Radford, Jong~Wook Kim, Chris Hallacy, Aditya Ramesh, Gabriel Goh, Sandhini Agarwal, Girish Sastry, Amanda Askell, Pamela Mishkin, Jack Clark, et~al.
\newblock Learning transferable visual models from natural language supervision.
\newblock In \emph{International conference on machine learning}, pages 8748--8763. PmLR, 2021.

\bibitem[Ruiz et~al.(2023)Ruiz, Li, Jampani, Pritch, Rubinstein, and Aberman]{ruiz2023dreambooth}
Nataniel Ruiz, Yuanzhen Li, Varun Jampani, Yael Pritch, Michael Rubinstein, and Kfir Aberman.
\newblock Dreambooth: Fine tuning text-to-image diffusion models for subject-driven generation.
\newblock In \emph{2023 IEEE/CVF Conference on Computer Vision and Pattern Recognition (CVPR)}, pages 22500--22510. IEEE, 2023.

\bibitem[She et~al.(2026)She, Fu, Liu, Jin, Wang, Liu, and Jiang]{she2026mosaic}
Dong She, Siming Fu, Mushui Liu, Qiaoqiao Jin, Hualiang Wang, Mu Liu, and Jidong Jiang.
\newblock Mosaic: Multi-subject personalized generation via correspondence-aware alignment and disentanglement.
\newblock In \emph{International Conference on Learning Representations}, pages 63128--63156, 2026.

\bibitem[Tian et~al.(2026)Tian, Mao, Liu, Zhang, Liu, Wu, and Wang]{tian2026multibind}
Wenqing Tian, Hanyi Mao, Zhaocheng Liu, Lihua Zhang, Qiang Liu, Jian Wu, and Liang Wang.
\newblock Multibind: A benchmark for attribute misbinding in multi-subject generation.
\newblock \emph{arXiv preprint arXiv:2603.21937}, 2026.

\bibitem[Wang et~al.(2024)Wang, Bai, Wang, Qin, Chen, Li, Tang, and Hu]{wang2024instantid}
Qixun Wang, Xu Bai, Haofan Wang, Zekui Qin, Anthony Chen, Huaxia Li, Xu Tang, and Yao Hu.
\newblock Instantid: Zero-shot identity-preserving generation in seconds.
\newblock \emph{arXiv preprint arXiv:2401.07519}, 2024.

\bibitem[Wang et~al.(2026)Wang, Wei, He, Ouyang, Lu, Zhao, and Tian]{wang2026psr}
Shulei Wang, Longhui Wei, Xin He, Jianbo Ouyang, Hui Lu, Zhou Zhao, and Qi Tian.
\newblock Psr: Scaling multi-subject personalized image generation with pairwise subject-consistency rewards.
\newblock In \emph{Proceedings of the IEEE/CVF Conference on Computer Vision and Pattern Recognition}, pages 14756--14766, 2026.

\bibitem[Wang et~al.(2025)Wang, Fu, Huang, He, and Jiang]{wang2025ms}
Xierui Wang, Siming Fu, Qihan Huang, Wanggui He, and Hao Jiang.
\newblock Ms-diffusion: Multi-subject zero-shot image personalization with layout guidance.
\newblock In \emph{International Conference on Learning Representations}, pages 95118--95146, 2025.

\bibitem[Wang et~al.(2003)Wang, Simoncelli, and Bovik]{wang2003multiscale}
Zhou Wang, Eero~P Simoncelli, and Alan~C Bovik.
\newblock Multiscale structural similarity for image quality assessment.
\newblock In \emph{The thrity-seventh asilomar conference on signals, systems \& computers, 2003}, pages 1398--1402. IEEE, 2003.

\bibitem[Wang et~al.(2004)Wang, Bovik, Sheikh, and Simoncelli]{wang2004image}
Zhou Wang, Alan~C Bovik, Hamid~R Sheikh, and Eero~P Simoncelli.
\newblock Image quality assessment: from error visibility to structural similarity.
\newblock \emph{IEEE transactions on image processing}, 13\penalty0 (4):\penalty0 600--612, 2004.

\bibitem[Wu et~al.(2025)Wu, Zheng, Yan, Xiao, Luo, Wang, Li, Jiang, Liu, Zhou, et~al.]{wu2025omnigen2}
Chenyuan Wu, Pengfei Zheng, Ruiran Yan, Shitao Xiao, Xin Luo, Yueze Wang, Wanli Li, Xiyan Jiang, Yexin Liu, Junjie Zhou, et~al.
\newblock Omnigen2: Exploration to advanced multimodal generation.
\newblock \emph{arXiv preprint arXiv:2506.18871}, 2025.

\bibitem[Wu et~al.(2023)Wu, Hao, Sun, Chen, Zhu, Zhao, and Li]{wu2023human}
Xiaoshi Wu, Yiming Hao, Keqiang Sun, Yixiong Chen, Feng Zhu, Rui Zhao, and Hongsheng Li.
\newblock Human preference score v2: A solid benchmark for evaluating human preferences of text-to-image synthesis.
\newblock \emph{arXiv preprint arXiv:2306.09341}, 2023.

\bibitem[Xiao et~al.(2025)Xiao, Yin, Freeman, Durand, and Han]{xiao2025fastcomposer}
Guangxuan Xiao, Tianwei Yin, William~T Freeman, Fr{\'e}do Durand, and Song Han.
\newblock Fastcomposer: Tuning-free multi-subject image generation with localized attention.
\newblock \emph{International Journal of Computer Vision}, 133\penalty0 (3):\penalty0 1175--1194, 2025.

\bibitem[Xu et~al.(2026)Xu, Cheng, Xing, Fang, Wu, Wang, Zeng, Jiang, Yu, Ma, et~al.]{xu2026withanyone}
Hengyuan Xu, Wei Cheng, Peng Xing, Yixiao Fang, Shuhan Wu, Rui Wang, Xianfang Zeng, Daxin Jiang, Gang Yu, Xingjun Ma, et~al.
\newblock Withanyone: Toward controllable and id consistent image generation.
\newblock In \emph{International Conference on Learning Representations}, pages 130581--130610, 2026.

\bibitem[Xu et~al.(2023)Xu, Liu, Wu, Tong, Li, Ding, Tang, and Dong]{xu2023imagereward}
Jiazheng Xu, Xiao Liu, Yuchen Wu, Yuxuan Tong, Qinkai Li, Ming Ding, Jie Tang, and Yuxiao Dong.
\newblock Imagereward: Learning and evaluating human preferences for text-to-image generation.
\newblock \emph{Advances in Neural Information Processing Systems}, 36:\penalty0 15903--15935, 2023.

\bibitem[Xu et~al.(2025)Xu, Wang, Peng, and Liu]{xu2025interact}
Zhu Xu, Zhaowen Wang, Yuxin Peng, and Yang Liu.
\newblock Interact-custom: Customized human object interaction image generation.
\newblock In \emph{Proceedings of the 33rd ACM International Conference on Multimedia}, pages 9500--9508, 2025.

\bibitem[Ye et~al.(2023)Ye, Zhang, Liu, Han, and Yang]{ye2023ip}
Hu Ye, Jun Zhang, Sibo Liu, Xiao Han, and Wei Yang.
\newblock Ip-adapter: Text compatible image prompt adapter for text-to-image diffusion models.
\newblock \emph{arXiv preprint arXiv:2308.06721}, 2023.

\bibitem[{Z.ai Team}(2026)]{zai2026glmimage}
{Z.ai Team}.
\newblock {GLM-Image}: Auto-regressive for dense-knowledge and high-fidelity image generation.
\newblock \url{https://z.ai/blog/glm-image}, 2026.

\bibitem[Zhai et~al.(2023)Zhai, Mustafa, Kolesnikov, and Beyer]{zhai2023sigmoid}
Xiaohua Zhai, Basil Mustafa, Alexander Kolesnikov, and Lucas Beyer.
\newblock Sigmoid loss for language image pre-training.
\newblock In \emph{2023 IEEE/CVF International Conference on Computer Vision (ICCV)}, pages 11941--11952. IEEE, 2023.

\end{thebibliography}
}
\onecolumn
\appendix
\clearpage
\setcounter{page}{1}
{\centering\Large\textbf{\thetitle}\\
\vspace{0.5em}Supplementary Material\\
\vspace{1.0em}\par}

\section{Dataset Statistics and Prompt Construction}
\begin{table}[h]
\centering
\small
\renewcommand{\arraystretch}{1.25}
\setlength{\tabcolsep}{4pt}
\scalebox{0.8}{
\begin{tabular}{p{0.23\linewidth}p{0.33\linewidth}p{0.35\linewidth}}
\toprule
\textbf{Attribute} & \textbf{Silver Set} & \textbf{Gold Set} \\
\midrule
Purpose 
& MIE training 
& Human benchmark evaluation \\
Scale 
& 60,000 candidate pairs; 56,909 dual-judge consensus pairs plus 146 single-judge records 
& 4,020 human-labeled pairs (3,690 with consensus) \\
Prompt coverage 
& Hierarchical prompts with nested complexity subsets, approximately uniform across relation type, human/object ratio, and subject count. & Nested prompts for unseen generators; a separate stratified prompt draw for seen generators. \\

Candidate generators 
& Nano Banana, MOSAIC 
& GPT-Image-1.5, Flux.2, Seedream 4.5, GLM; plus Nano Banana/MOSAIC for seen-generator evaluation \\
Annotation source 
& SOP-guided dual-VLM consensus, plus 146 single-judge records 
& Double-blind human annotation \\
Annotators 
& Gemini-2.5-Flash and Gemini-3.1-Flash-Lite 
& Two independent human annotators per pair \\
Label types 
& Pairwise preference; Existence; Appearance; Interaction 
& Pairwise preference; Existence; Appearance; Interaction \\
Evaluation role 
& Train and Val
& Seen-generator: 1,500 pairs; unseen-generator: 2,520 pairs \\
\bottomrule
\end{tabular}
}
\caption{Dataset statistics for MIB. The Silver Set provides scalable SOP-guided VLM supervision for training MIE, while the Gold Set is fully human-labeled under a double-blind protocol.}
\label{tab:dataset_statistics}
\end{table}

\begin{figure}[htpb]
    \centering
    \includegraphics[
        width=\linewidth
    ,height=0.6\textheight,keepaspectratio]{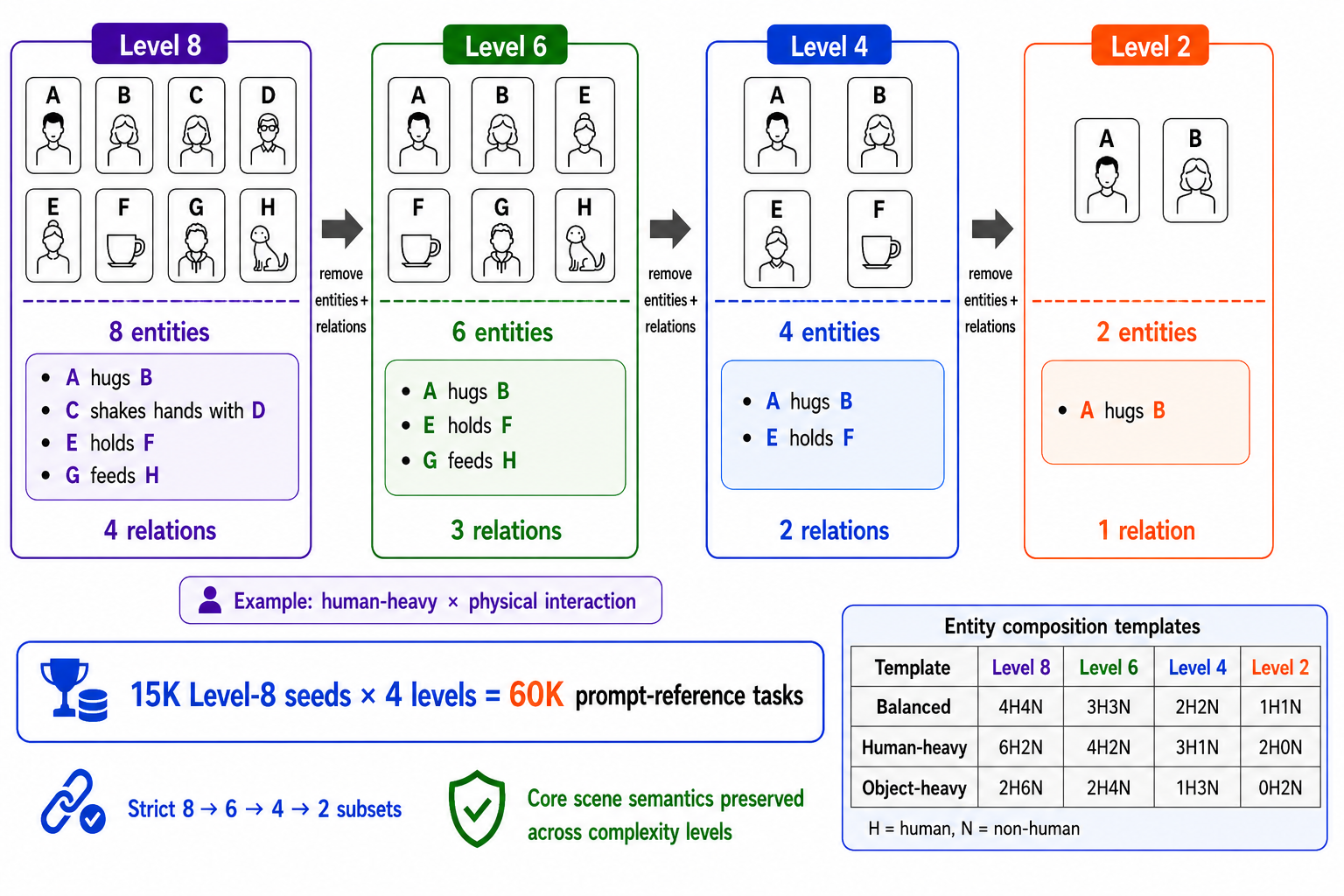}
    \caption{
    Hierarchical prompt construction and entity-composition templates. 
    Each Level-8 seed is reduced into strict Level-6/4/2 subsets by removing entities together with their associated relations, preserving core scene semantics while varying entity-relation density. 
    The example illustrates a human-heavy physical-interaction bucket, while the table summarizes the human/non-human composition templates used across balanced, human-heavy, and object-heavy ratio types.
    }
    \label{fig:hierarchical_prompt_construction}
\end{figure}

\subsection{Overlap between MIB-Silver and MIB-Gold}
\label{sec:app_overlap}
Gold shares its reference identities with Silver by design: all 80 identities of the unseen-generator split and 79 of the 80 in the seen-generator split occur in Silver, which uses a closed pool with one reference image per identity. The unseen-generator split re-renders 1{,}260 Silver prompts with four generators that never appear in Silver. The seen-generator split is a separate prompt draw: none of its 1{,}488 distinct prompts matches a Silver prompt literally, and 45 match after normalizing case, whitespace, and one fixed sentence. Of the 60{,}000 Silver tasks, 2{,}945 were dropped before splitting (2{,}943 because the two judges chose different winners). The remaining 57{,}055 are split by seed group into train / val / test (51{,}328 / 2{,}847 / 2{,}880), with no seed and no prompt in two splits. MIE is trained for 600 optimizer steps with batch size 16 (9{,}600 pairs) drawn from the training split only.

Table~\ref{tab:overlap} partitions the 2{,}279 consensus pairs of the unseen-generator split by where their prompt lies in Silver. Row~1 records split membership, not exposure, since only 9{,}600 of the 51{,}328 training pairs were used; identities are shared in every row, so identity familiarity is tested only by the reference-extension study (Sec.~\ref{appendix:robust}).

\begin{table}[h]
\centering
\small
\caption{MIE accuracy on the unseen-generator split by the Silver split of each prompt (95\% bootstrap CIs over prompt-seed groups). Rows 1--3 partition the 2{,}279 consensus pairs.}
\label{tab:overlap}
\begin{tabular}{@{}lrc@{}}
\toprule
Unseen-split pairs & $n$ & MIE acc. (95\% CI) \\
\midrule
1. Prompt in the training split & 1,883 & 0.884 (0.870--0.898) \\
2. Prompt from a val/test seed group & 286 & 0.895 (0.861--0.928) \\
3. Prompt absent from released Silver splits & 110 & 0.855 (0.787--0.917) \\
Rows 2+3: prompt outside the training split & 396 & 0.884 (0.853--0.914) \\
\bottomrule
\end{tabular}
\end{table}

\FloatBarrier
\section{Appendix: Human Annotation Observations and Failure Mode Analysis}
\label{sec:appendix_failure_modes}
All reference images used for human annotation are standardized as isolated single-subject studio images at $512 \times 512$, allowing the same identity anchors to be recombined across prompts. For each prompt-reference task, a personalized generator receives the guiding prompt and reference images as visual conditioning inputs, producing a candidate image. MIB forms pairwise comparisons by placing two candidates under the same prompt-reference condition, so labels capture relative generation quality rather than prompt difficulty. References are generated separately from all candidate generators to reduce generator-specific leakage.

Table~\ref{tab:similarity_cooccurrence} reports the full breakdown across ethnicity, age band, and wardrobe (gray T-shirt, jeans), for MIB-Silver and MIB-Gold overall and by subject count $N$ in MIB-Silver; co-occurrence of similar subjects increases from $N=2$ to $N=8$ across every axis. This design directly targets our Appearance dimension: when two visually or demographically similar subjects share a scene, generators that rely on coarse cues such as ethnicity, hairstyle, or clothing rather than fine-grained facial identity are exposed, and an identity mix-up is scored as an Appearance failure (wrong face, hair, or garment attributed to the wrong subject, including cross-subject feature bleeding; Appendix~\ref{sec:appendix_failure_modes}), while Existence separately catches the more severe case where two similar subjects collapse into one. 

\begin{table}[H] \centering \caption{Co-occurrence rate of visually or demographically similar human subjects within a scene, overall in MIB-Silver and MIB-Gold, and by subject count $N$ in MIB-Silver.} \label{tab:similarity_cooccurrence} \begin{tabular}{@{}lcccccc@{}} \toprule Metric & Silver & Gold & $N=2$ & $N=4$ & $N=6$ & $N=8$ \\ \midrule Same ethnicity & 39.4\% & 38.9\% & 7.3\% & 26.4\% & 54.7\% & 69.0\% \\ Same age band & 47.4\% & 47.1\% & 10.8\% & 35.3\% & 67.4\% & 76.3\% \\ Both gray T-shirt & 16.9\% & 16.9\% & 2.6\% & 9.5\% & 21.0\% & 34.4\% \\ Both jeans & 41.7\% & 41.3\% & 10.4\% & 30.3\% & 57.2\% & 69.0\% \\ \bottomrule \end{tabular} \end{table} 

During the human annotation process for the gold test set, annotators recorded recurring failure patterns across evaluated image pairs. These observations surfaced three consistent and practically significant findings, which we document here to inform future dataset construction and model development efforts.

\paragraph{Failure Rates Increase with Subject Count.}
Annotators consistently found that images depicting a larger number of subjects were harder to evaluate correctly, and that generative models performed more poorly on such cases. The most common failure modes in high-subject-count scenes were subject omission, where one or more individuals were missing from the generated output, and appearance drift, where the physical attributes of individual subjects degraded as the total number of subjects increased. This trend was particularly pronounced for visually similar subjects (e.g., individuals wearing comparable outfits or sharing similar physical features), where models frequently confused or merged identities. Facial attributes and clothing details were identified as the two most error-prone visual channels under these conditions.


\begin{figure}[hbpt]
  \centering
    \includegraphics[width=0.75\textwidth,height=0.6\textheight,keepaspectratio]{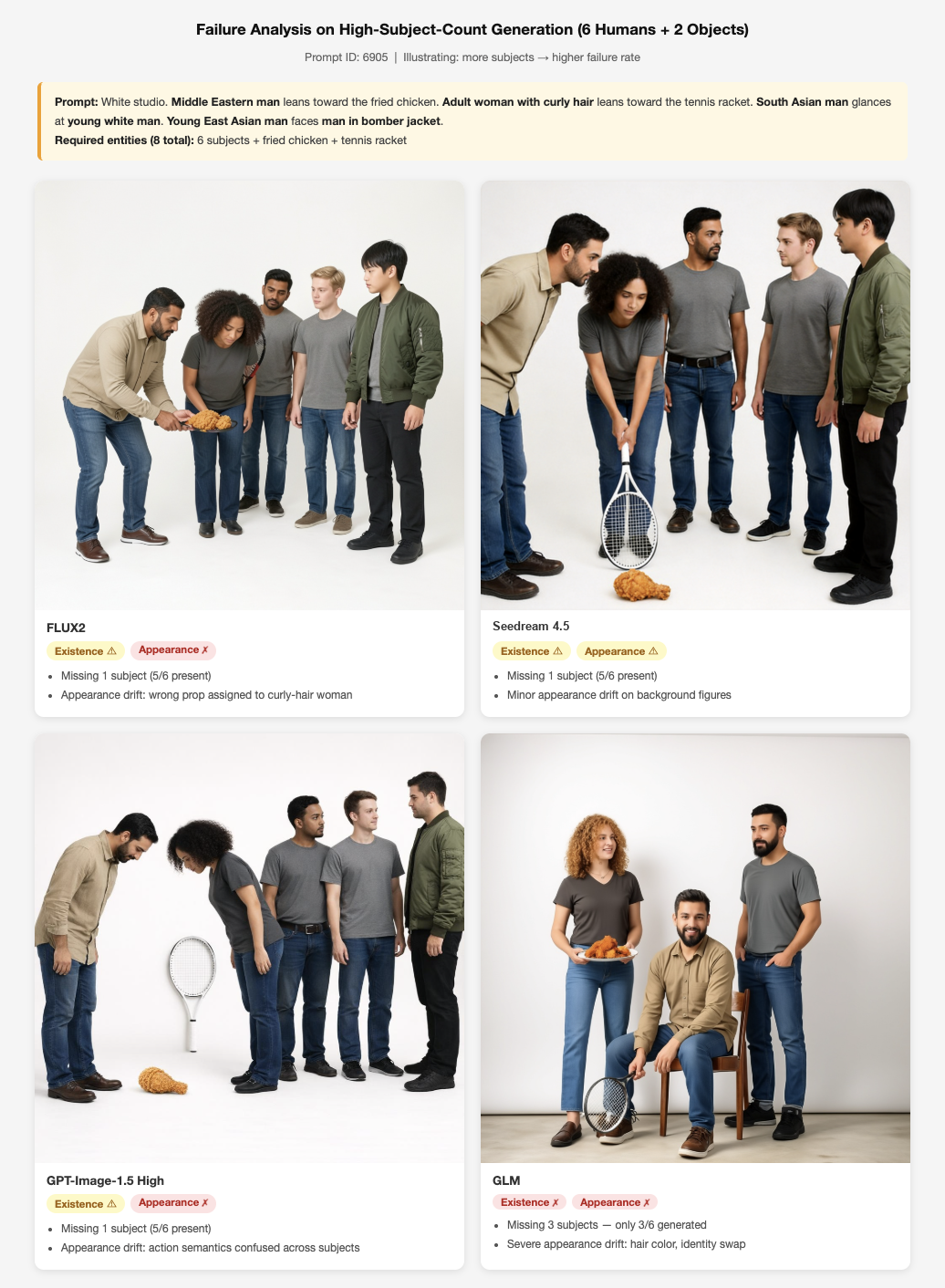}
  \caption{\textbf{Failure Rates Increase with Subject Count} A representative 6-subject, 2-object prompt (ID: 6905) evaluated across four generative models. All models exhibit Existence failures, with subject omission ranging from 1 missing subject (FLUX2, Seedream-4.5, GPT) to 3 missing subjects (GLM).}
  \label{fig:appendix-failure-rates-subject-count}
\end{figure}

\paragraph{Existence Failures Frequently Co-occur with Appearance Failures.}
A substantial proportion of annotated samples exhibited a systematic co-occurrence between Existence and Appearance failures. When a model fails to render all requested subjects as distinct entities, it often compensates by blending the features of missing subjects into the remaining ones, for instance, combining the facial identity of subject $A$ with the wardrobe of subject $B$ into a single generated figure. This entanglement means that Existence failures are rarely isolated: they tend to cascade into cross-subject feature bleeding, inflating Appearance failure rates in a correlated rather than independent manner. This observation motivates treating the three diagnostic dimensions as structured but not fully orthogonal signals in practice, despite their conceptual independence.


\begin{figure}[htbp]
  \centering
  \includegraphics[width=0.8\textwidth,height=0.6\textheight,keepaspectratio]{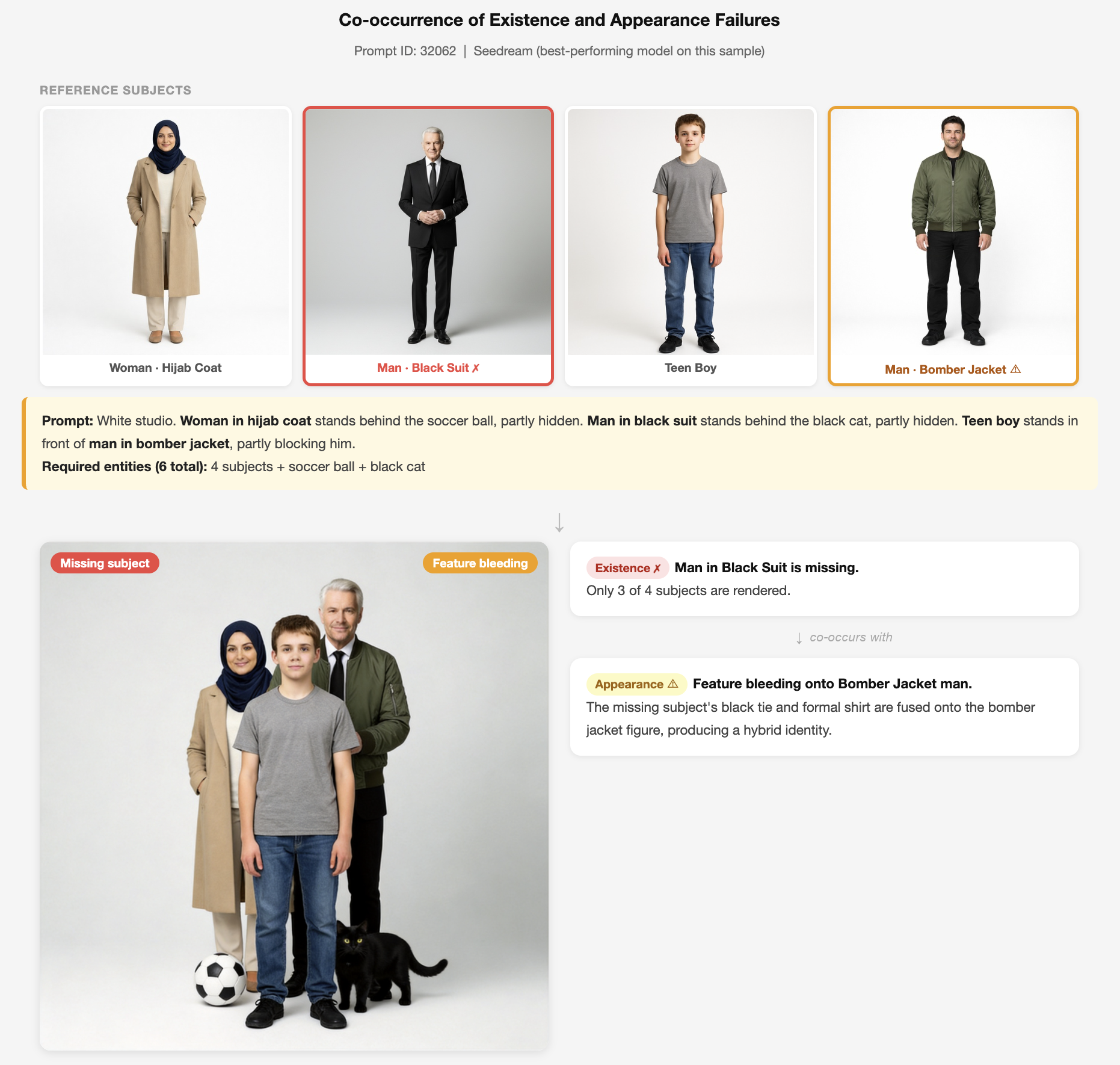}
  \caption{\textbf{Existence Failures Frequently Co-occur with Appearance Failures} A representative six-subject prompt (ID: 32062) generated by Seedream, the human-preferred model in this pair. The missing subject (Man in Black Suit) is not simply absent: its visual attributes,  the black tie and formal shirt,  are redistributed onto the surviving Bomber Jacket figure, producing a hybrid identity.}
  \label{fig:appendix-existence-appearance-cooccurrence}
\end{figure}

\begin{figure}[htbp]
  \centering
  \includegraphics[width=0.7\textwidth,height=0.6\textheight,keepaspectratio]{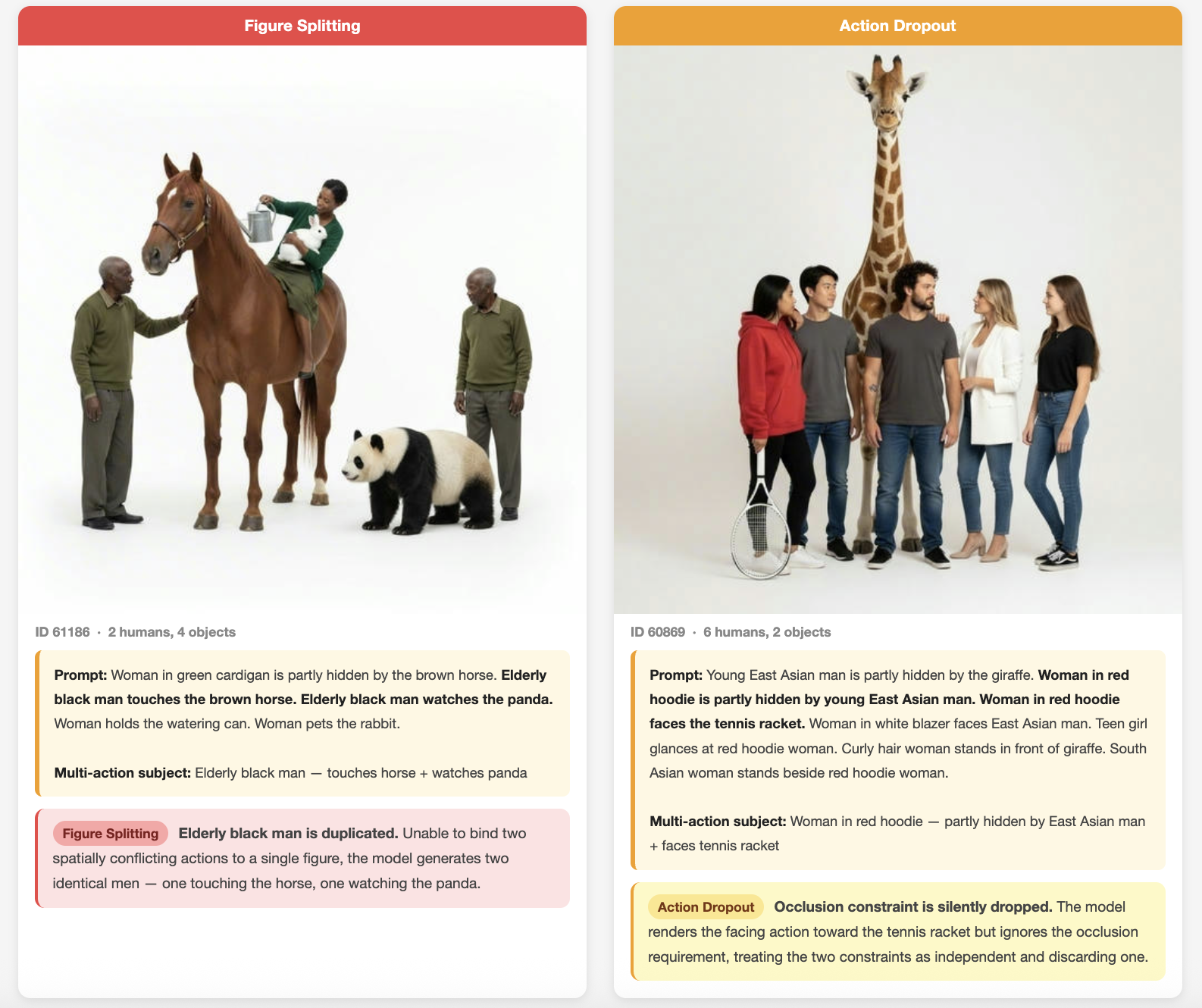}
  \caption{\textbf{Illustrative Multi-Action Failures.} These two Nano Banana examples show a subject with more than one concurrent action either cloned to satisfy each action separately (Figure Splitting, ID: 61186) or rendered with one of its assigned actions dropped (Action Dropout, ID: 60869). Nano Banana was preferred over MOSAIC in both comparisons.}
  \label{fig:appendix-multi-action-overload}
\end{figure}

\paragraph{Illustrative Multi-Action Failures}
Annotators observed cases in which a single subject assigned more than one concurrent action or interaction was not rendered with all specified behaviors. We illustrate two failure patterns. First, \textit{figure splitting}: the model resolves the conflict by duplicating or fragmenting the subject, distributing distinct actions across two visually similar but non-identical figures,  effectively cloning the identity to satisfy each action independently. Second, \textit{action dropout}: the model selectively ignores or partially executes one of the assigned actions, producing a subject that satisfies only a subset of the prompt specification. Figure~\ref{fig:appendix-multi-action-overload} illustrates both patterns in Nano Banana.

\subsection{Structured Example Grid}
\label{sec:app_example_grid}

To make the compositional design of \texttt{MIB} concrete, Table~\ref{tab:structured_example_grid} presents one hierarchical seed per relation category (no interaction/no occlusion, occlusion without interaction, and physical interaction), showing its Level-8 prompt and the Level-2 subset derived from it. Lower levels keep the clauses of the higher level and remove entity pairs together with their relations.

\begin{table*}[h]
\centering
\small
\begin{tabular}{@{}p{0.16\linewidth} p{0.40\linewidth} p{0.40\linewidth}@{}}
\toprule
Interaction Class & 2 Subjects & 8 Subjects \\
\midrule
No Interaction \& No Occlusion &
\textit{``White studio. man denim jacket glances at the white mug.''} &
\textit{``White studio. man denim jacket glances at the white mug. young white woman looks at the dslr camera. middle eastern man faces the table lamp. south asian woman glances at the acoustic guitar.''} \\
\addlinespace
Occlusion without Interaction &
\textit{``White studio. The ice cream on the ground blocks part of south asian woman's feet.''} &
\textit{``White studio. The ice cream on the ground blocks part of south asian woman's feet. elderly black man is partly hidden by the rabbit on the ground. elderly east asian woman stands behind the red backpack on the ground, partly hidden. The sheep on the ground obscures young white man's lower legs.''} \\
\addlinespace
Physical Interaction &
\textit{``White studio. elderly black woman lifts the travel suitcase.''} &
\textit{``White studio. elderly black woman lifts the travel suitcase. adult woman curly hair holds the orange basketball. south asian man feeds the parrot. elderly black man feeds the koala.''} \\
\bottomrule
\end{tabular}
\caption{Level-2 and Level-8 prompts of three MIB-Silver seeds (IDs 20001, 25598, 33906), one per relation category. Each Level-2 prompt is the first clause of its Level-8 parent; the global ``Keep all entities at realistic scale'' sentence is omitted for brevity.}
\label{tab:structured_example_grid}
\end{table*}

Because each lower level is a strict subset of its seed, differences across levels come from adding subjects and relations rather than from changing the scene description.

\clearpage
\section{Appendix: Summarized Table of Related Work}
\label{sec:appendix_related_work}
Table~\ref{tab:benchmark_comparison} extends the comparison in Section~2 to additional benchmarks. Two properties distinguish MIB from the listed work: factorized relation-level stress and human pairwise gold labels.

\begin{table}[!h]
\centering
\footnotesize
\setlength{\tabcolsep}{2.5pt}
\renewcommand{\arraystretch}{1.0}
\begin{tabular}{@{}>{\raggedright}p{2.5cm} c >{\centering}p{0.9cm} >{\centering}p{1.1cm} >{\raggedright}p{3.1cm} >{\centering}p{1.4cm} >{\centering}p{1.9cm} >{\raggedright\arraybackslash}p{3.6cm}@{}}
\toprule
\textbf{Benchmark} 
& \textbf{Year}
& \textbf{Ref. Images} 
& \textbf{Multi-subj.} 
& \textbf{Stress / Eval. Design} 
& \textbf{Human Pairwise Gold}
& \textbf{Pref./Diag. Supervision}
& \textbf{Main Gap} \\
\midrule

T2I-CompBench~\citep{huang2023t2i}
& 2023
& \xmark
& Text-only
& Text-only composition
& \xmark$^{\dagger}$
& \xmark
& Cannot measure reference identity, identity collapse, or appearance leakage. \\

DreamBench~\citep{ruiz2023dreambooth}
& 2023
& \cmark
& \xmark
& Single-subject prompts
& \xmark
& \xmark
& No multi-subject binding or interaction evaluation. \\

DreamBench++\newline\citep{peng2025dreambench}
& 2025
& \cmark
& \xmark
& Single-subject, human-aligned automated eval.
& \xmark$^{\dagger}$
& \xmark
& Broader single-subject evaluation, but no high-subject-count binding stress. \\

CustomConcept101\newline\citep{kumari2023multi}
& 2023
& \cmark
& Limited
& Multi-concept composition
& \xmark
& \xmark
& Not designed for contact-rich or occlusion-driven identity binding. \\

XVerseBench~\citep{chen2025xverse}
& 2025
& \cmark
& \cmark
& Single / dual / triple subjects
& \xmark
& \xmark
& Limited high-subject-count and relation-level stress factorization. \\

PSRBench~\citep{wang2026psr}
& 2026
& \cmark
& \cmark
& Broad capability subsets
& \xmark
& Reward metrics
& Capability-oriented; not a factorized diagnostic benchmark for interaction/occlusion binding failures. \\

OmniContext~\citep{wu2025omnigen2}
& 2025
& \cmark
& \cmark
& In-context character / object / scene eval.
& \xmark
& Automated judge scores
& Broad in-context consistency benchmark, but lacks human gold and factorized relation-level binding stress. \\

MultiBanana~\citep{oshima2026multibanana}
& 2026
& \cmark
& \cmark
& Multi-reference challenges
& \xmark$^{\dagger}$
& VLM-based scores
& Covers multi-reference difficulty, but not hierarchical interaction/occlusion diagnosis or human pairwise gold labels. \\

MultiHuman-Testbench~\citep{borse2025multihuman}
& 2025
& \cmark
& \cmark
& Multi-human actions / faces
& \xmark
& Metric suite
& Restricted to human subjects; does not cover human-object or object-object binding. \\

MultiBind~\citep{tian2026multibind}
& 2026
& \cmark
& \cmark
& Attribute / slot misbinding
& \xmark$^{\dagger}$
& Diagnostic metrics
& Focuses on attribute-level misbinding rather than factorized physical interaction and occlusion stress. \\ MultiID-Bench (WithAnyone)~\citep{xu2026withanyone} & 2026 & \cmark & \cmark & Human ID-consistency \& pose/expression variation & \xmark & Face/identity metrics & Centered on human/character identity fidelity; no non-human subjects or occlusion/relation stress. \\

\midrule
\textbf{MIB (Ours)}
& \textbf{2026}
& \cmark
& \cmark
& \textbf{Factorized relation-level binding stress}
& \cmark
& \cmark
& -- \\
\bottomrule
\end{tabular}
\vspace{0.3em}
\caption{Comparison with existing benchmarks. DreamBench and CustomConcept101 refer to evaluation sets from DreamBooth and Custom Diffusion respectively. $^{\dagger}$Reports other forms of human evaluation (e.g., ratings or validation studies) but no pairwise human gold labels for cross-generator ranking. MIB uniquely targets factorized relation-level binding stress with both scalable silver supervision and human gold evaluation.}
\label{tab:benchmark_comparison}
\end{table}

\FloatBarrier
\section{Appendix: MIB Reference Subjects}
\label{appendix:reference_pool}
MIB uses a fixed pool of 80 GPT-Image-1 reference subjects, including 30 human and 50 non-human identities. Human references vary in age, gender presentation, ethnicity, clothing, hairstyle, and appearance, explicitly spanning a wide range of ages (teens to elderly), diverse ethnicities, unique personal attributes (e.g., curly, braided, or bald hairstyles), and specific cultural attire (e.g., hijabs or distinctive coats); to introduce deliberate similarity, half of the subjects share near-identical casual wear (gray T-shirts and jeans), so identity cannot simply be read off clothing, and visually similar subjects further co-occur within prompts, forcing models to preserve fine-grained facial features rather than relying on attire shortcuts. The 50 non-human references cover long-tailed classes spanning animals, foods, tools, and objects, including occluders. Table~
\ref{tab:reference_distribution} summarizes the demographic and category composition of the 80-subject reference pool. 

\begin{table}[h] \centering \small \caption{Demographic and category distribution of the 80-subject reference pool (30 human, 50 non-human). All human identities are synthetic; age and ethnicity are as depicted, annotated by the authors.} \label{tab:reference_distribution} \begin{tabular}{@{}p{0.26\linewidth}p{0.64\linewidth}@{}} \toprule \multicolumn{2}{l}{ \textit{Humans (30)}} \\ \midrule Gender & 15 male / 15 female \\ Age band & 2 teens, 6 young adults (18--30), 15 adults (30--60), 7 elderly (60+) \\ Ethnicity (as depicted) & 11 White, 7 Black, 6 East Asian, 3 South Asian, 3 Middle Eastern \\ Hair style & 15 short, 5 long straight, 3 bob/shoulder-length, 3 curly/coiled, 1 braided, 1 bald, 2 hijab-covered \\ Hair color & $\sim$15 black/dark, $\sim$6 brown, 2 light/blond, 7 gray/white \\ Facial features & 3 with facial hair, 2 hijab, 1 bald \\ Tops & 9 gray T-shirts, 3 gray sweaters, 18 distinctive tops (flannel, hoodie, blazer, suit, dress, coat, cardigan, colored tees) \\ Bottoms & 17 jeans, 6 dark trousers, 3 khaki chinos, 4 other (joggers/skirt/dress/cream trousers) \\ \addlinespace \multicolumn{2}{l}{ \textit{Non-humans (50)}} \\ \midrule Animals (20) & 10 wild mammals, 6 domestic/farm, 4 birds \\ Foods (10) & 7 savory, 2 desserts, 1 fruit \\ Objects (20) & 5 sports equipment, 4 electronics/photography, 4 household, 6 travel/tools, 1 musical instrument \\ \bottomrule \end{tabular} \end{table} Seven references (4 large animals, 3 large objects) additionally serve as valid person-occluders in occlusion prompts, and the wardrobe overlap described above is realized concretely: 39.4\% of Silver scenes contain a same-ethnicity human pair, rising to 69.0\% at eight subjects. Two Middle Eastern subjects specifically (e.g., a \emph{middle eastern man} paired with a \emph{middle eastern woman in hijab}) co-occur in 2.2\% of Silver scenes, rising to 4.9\% at eight subjects, directly instantiating this same-ethnicity case.

We use synthetic, GPT-Image-1 generated references rather than real photographs for two reasons. First, releasing real human photographs at benchmark scale raises copyright and consent concerns that would compromise full reproducibility and public redistribution; synthetic references avoid this while following recent personalization benchmarks for fair comparability, and all synthetic images are kept strictly photorealistic so that identity preservation is not trivialized into an easy proxy task. Second, several stress conditions in MIB, such as same-colored clothing or matching hairstyles, require subjects that differ only along one targeted axis, combinations that are difficult to source from real photographs at scale but straightforward to construct precisely with generated references. Assembling an equivalent pool from licensed real photographs would also force acceptance of that corpus's own demographic and category imbalances, confounding the very axes MIB is designed to isolate. Generated references further provide a single canonical view per identity that stays fixed across every task instance in which that identity appears, giving Appearance fidelity a stable, neutral anchor.

\FloatBarrier
\section{Appendix: SOP Prompt}
\label{sec:app_SOP}
\begin{verbatim}
You are an expert judge for multi-subject personalized image generation.

You will receive:
1. The original generation prompt.
2. Reference images for each subject mentioned in the prompt.
3. Candidate image A.
4. Candidate image B.

Your job consists of two parts:
Part 1: Independent Evaluation
Evaluate candidate A and candidate B independently. For each image, 
strictly evaluate:
- Existence: Are ALL the required subjects in the references physically 
present?
  (Generic versions count as 1; completely missing counts as 0).
- Appearance: Are the subjects rendered WITHOUT structural deformations? 
  Do the generated subjects strictly match the references across 
  ALL features: Face/Skin tone, Hair style, Top clothing, AND Bottoms (pants)?
- Interaction: Does the generated image reflect the described actions, maintain
  correct proportions, and follow realistic spatial/physical laws?

Part 2: Overall Comparison
Compare Candidate A and B and decide the overall winner.

OUTPUT RULES:
- Use ONLY the provided prompt and images.
- All dimensional scores (`*_existence', `*_appearance', `*_interaction') must be
  strictly binary: 1 (fully correct) or 0 (any issue exists).
- `winner' must be either "A" or "B" (Never "Tie").
- Return valid JSON ONLY. Absolutely NO markdown formatting.
- CRITICAL: To focus on errors, the JSON MUST begin with `a_flaw_log` and
  `b_flaw_log`. DO NOT list perfect subjects. ONLY list subjects with errors and
  explicitly state what is wrong (Missing, Wrong Face/Hair/Skin, Wrong Top,
  Wrong Pants, Action/Spatial/Physics Error). If an image is 100% perfect,
  write "None".

SCORING NOTES & SPECIFIC EDGE CASES:
- Appearance (Strict Identity & Wardrobe): You MUST check the whole person. If a
  subject has the correct top but wrong pants, wrong hair, or mismatched skin
  tone/face compared to the reference, score Appearance as 0 and log the
  specific flaw (e.g., "Woman red hoodie: Wrong face/skin color",
  "Man flannel shirt: Wrong pants color").
- Appearance (Mangled/Melted): Any structural artifacts (extra limbs, mangled
  faces) = Appearance 0.
- Existence vs Appearance: If requested "man denim jacket" is drawn as a generic
  man in a grey sweater, Existence is 1, Appearance is 0.
- Interaction (Partial): Evaluate interactions based ONLY on rendered subjects.
  If an interaction requires a missing subject, Interaction is 0.
- Interaction (Physics, Environment & Scale): Obvious physical violations
  (e.g., objects floating in mid-air without support) OR severe scale distortions
  between objects (e.g., abnormally giant bread, miniature people) = Interaction 0.
  HOWEVER, the presence of unprompted but logical supporting structures
  (e.g., tables, pedestals) used to hold objects is PERMITTED and should NOT be
  penalized.

EXPECTED JSON SCHEMA:
{
  "a_flaw_log": "Middle eastern man: Appearance (wears green sweater instead of
  beige shirt). Man flannel shirt: Appearance (wrong pants color). Woman red
  hoodie: Appearance (mismatched face/skin/hair). Interaction: Man is not
  occluded by the woman; The burger is floating in mid-air.",
  "b_flaw_log": "Middle eastern man: Missing. Man bomber jacket: Missing. DSLR
  camera: Missing. Interaction: Man is not walking towards/staring at the woman;
  Fails due to missing interaction targets.",
  "a_existence": 1,
  "a_appearance": 0,
  "a_interaction": 0,
  "b_existence": 0,
  "b_appearance": 0,
  "b_interaction": 0,
  "winner": "A"
}
\end{verbatim}

\newpage
\FloatBarrier
\section{Appendix: Demos of Generated Training Labels}
\label{sec:app_SOP_demo}

\begin{figure}[!ht]
    \centering
    \begin{subfigure}[b]{0.48\textwidth}\includegraphics[width=\textwidth,height=0.4\textheight,keepaspectratio]{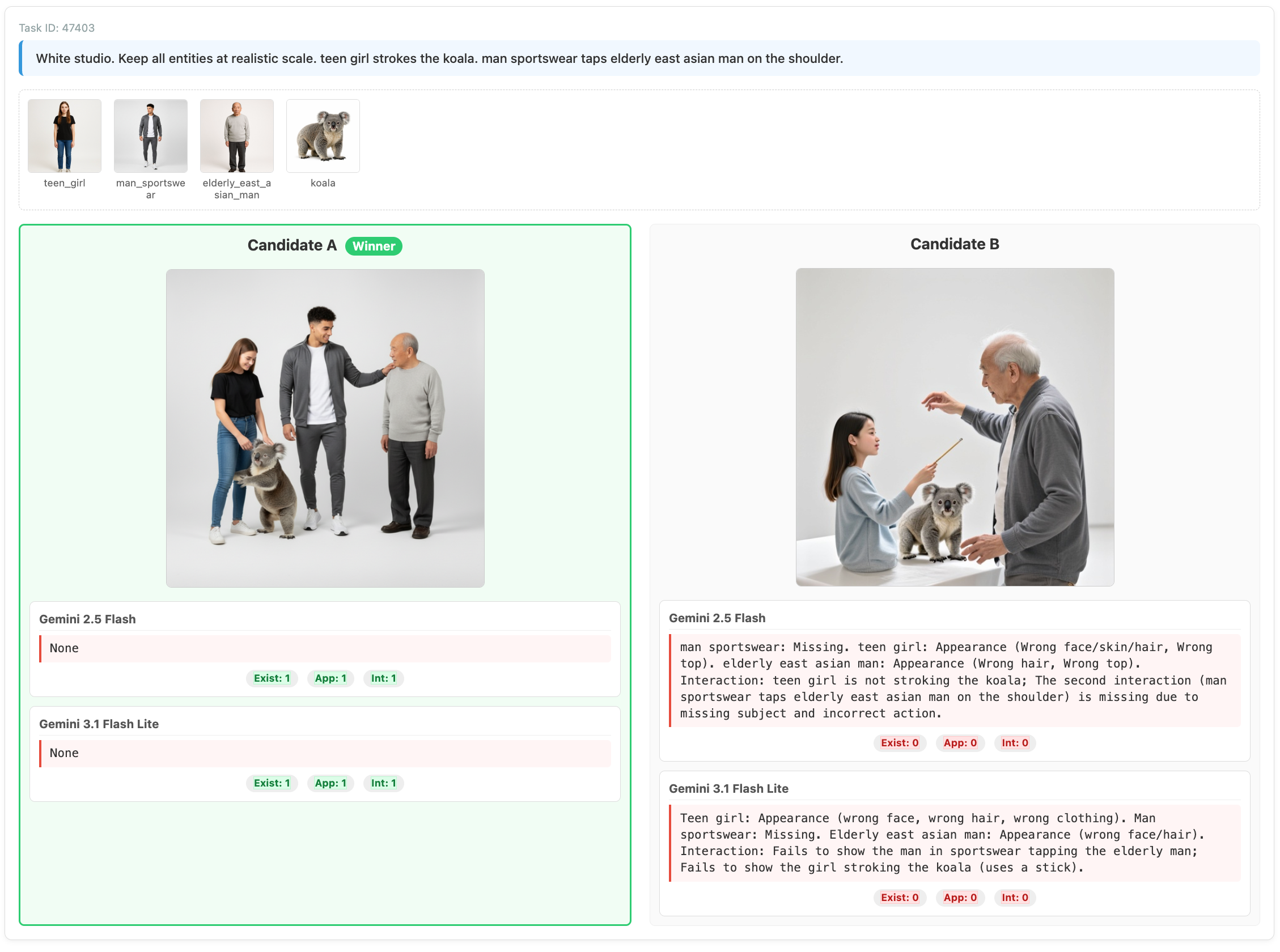}\caption{Action and target errors}\end{subfigure}\hfill
    \begin{subfigure}[b]{0.48\textwidth}\includegraphics[width=\textwidth,height=0.4\textheight,keepaspectratio]{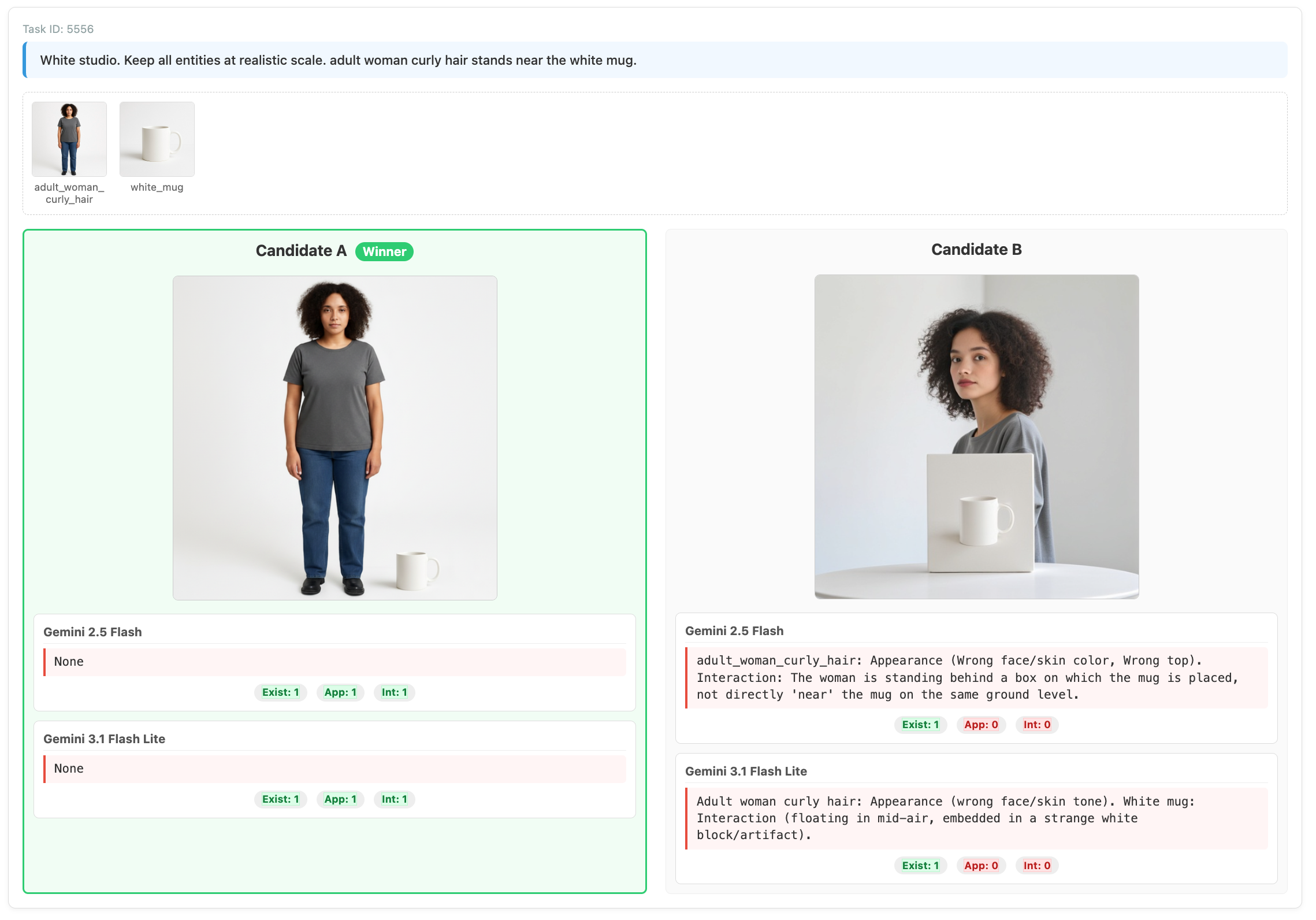}\caption{Physics and spatial violations}\end{subfigure}\\[0.8em]
    \begin{subfigure}[b]{0.48\textwidth}\includegraphics[width=\textwidth,height=0.4\textheight,keepaspectratio]{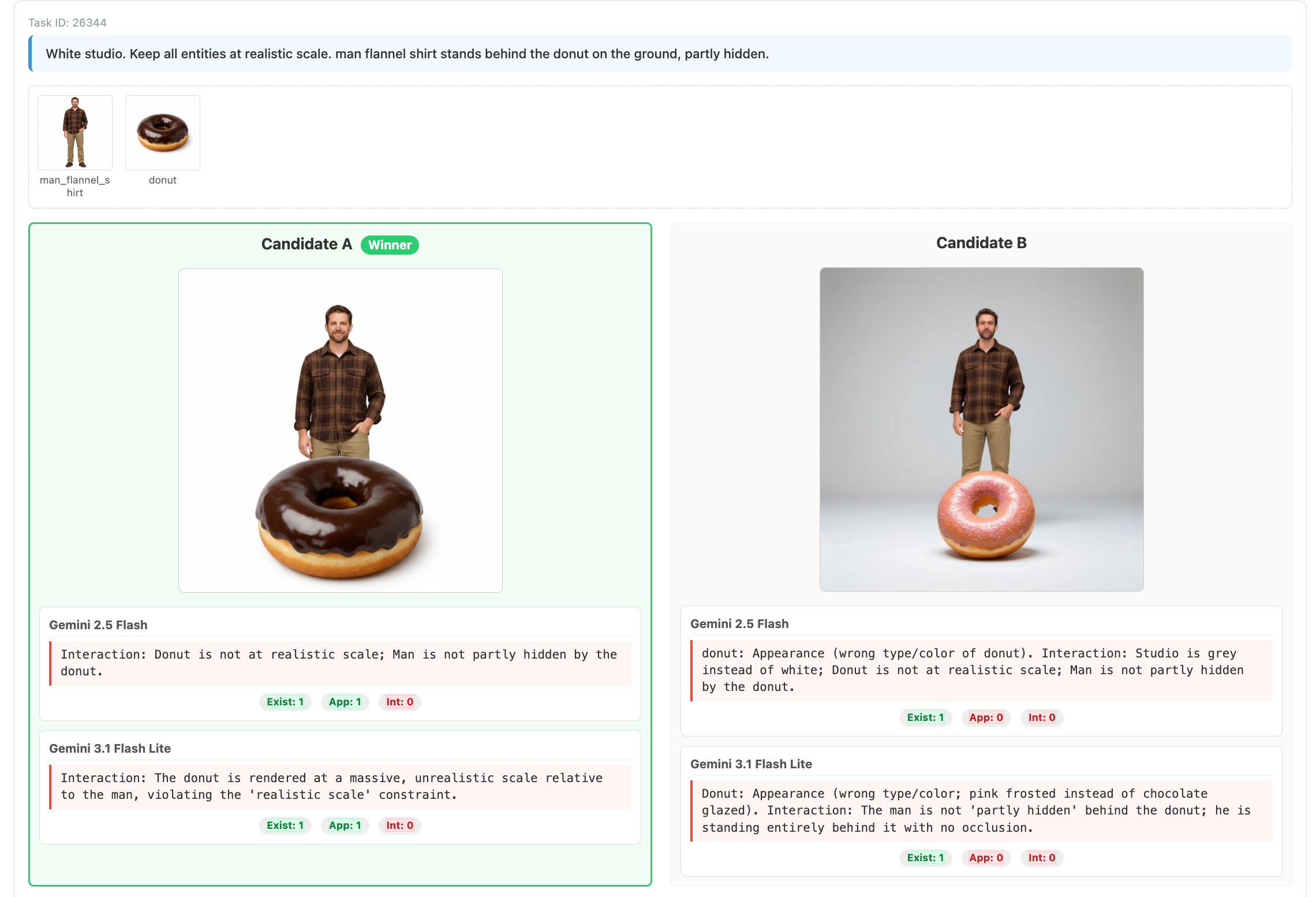}\caption{Appearance and occlusion errors}\end{subfigure}\hfill
    \begin{subfigure}[b]{0.48\textwidth}\includegraphics[width=\textwidth,height=0.4\textheight,keepaspectratio]{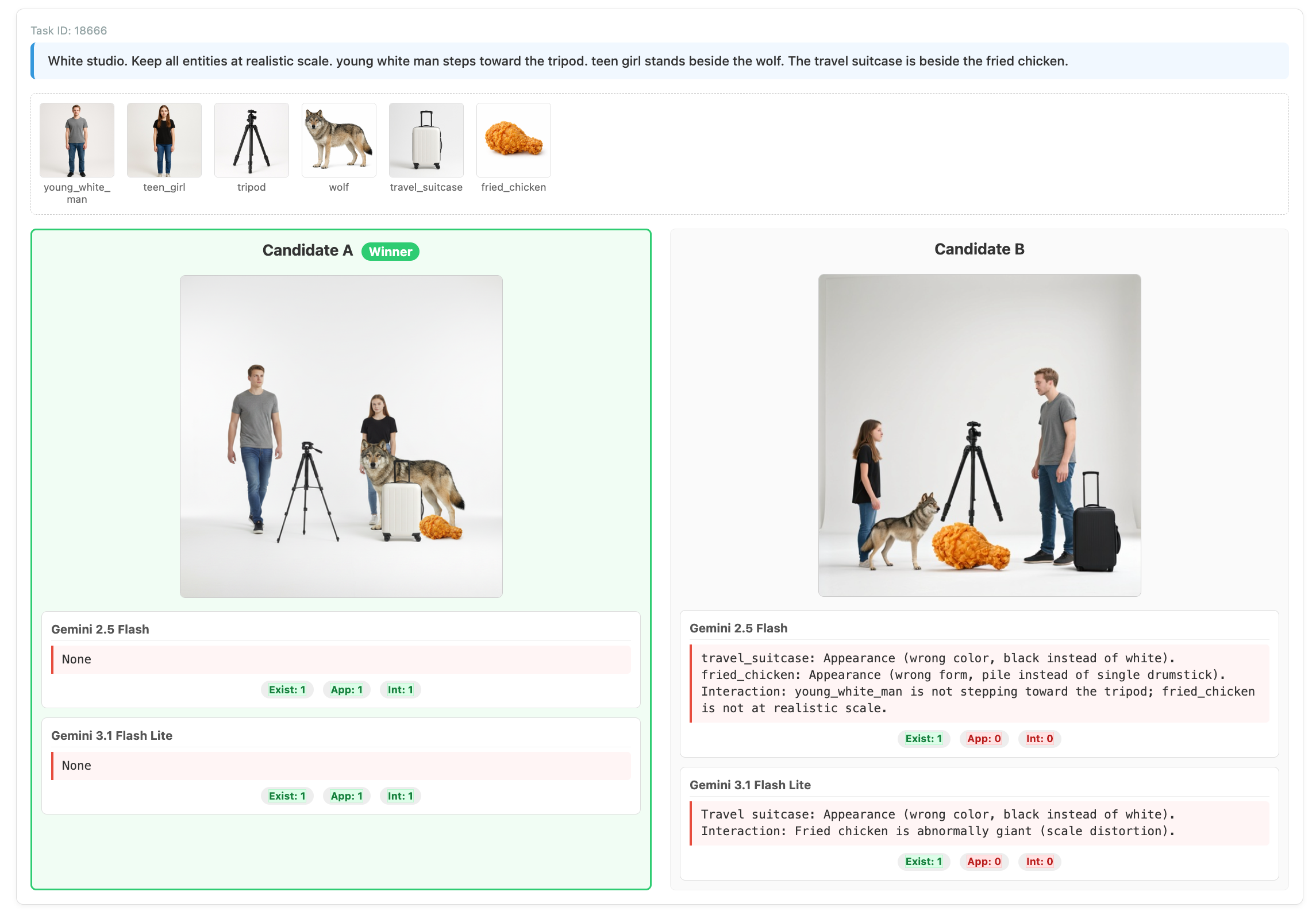}\caption{Scale distortion and color mismatch}\end{subfigure}
    \caption{Examples of Silver labels on which both VLM judges agree. The flaw logs show the SOP being applied to appearance, physical realism, and interaction errors.}
    \label{fig:sop_demos}
\end{figure}
\begin{figure}[p]
    \ContinuedFloat
    \centering
    \begin{subfigure}[b]{0.9\textwidth}\includegraphics[width=\textwidth,height=0.8\textheight,keepaspectratio]{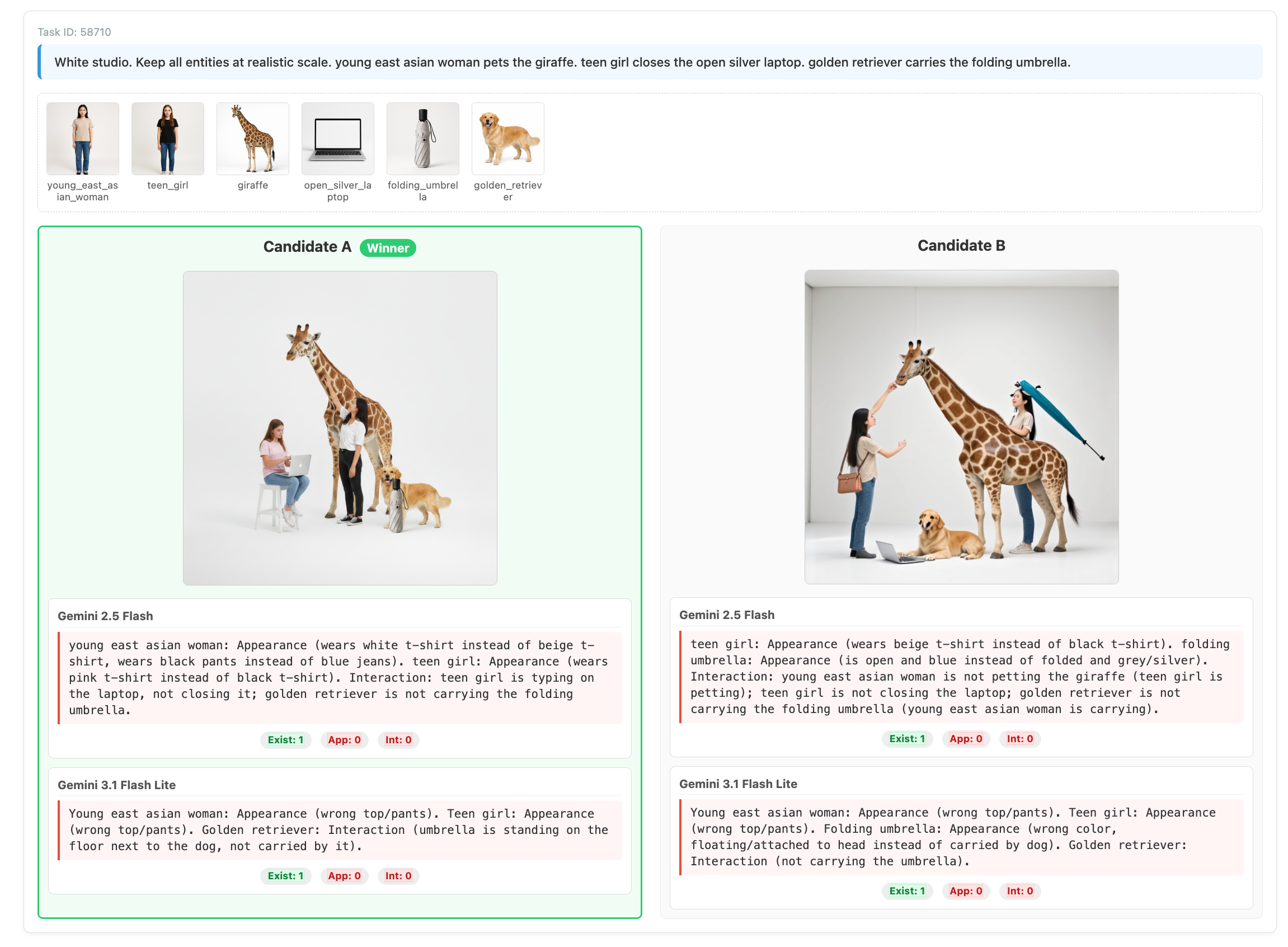}\caption{A six-subject scene}\end{subfigure}
    \caption{(continued) Silver labels for a six-subject scene.}
\end{figure}

\FloatBarrier
\section{MIE Configuration}
\label{sec:appendix_configuration}
\subsection{Diagnostic Validation}
\label{sec:app_diag}
Diagnostic F1 uses the 7{,}380 images in preference-consistent Gold pairs, retaining 6{,}841 / 5{,}988 / 5{,}565 images with agreeing labels for Existence / Appearance / Interaction. A post hoc threshold sweep at 0.3 / 0.4 / 0.5 / 0.6 / 0.7 gives macro-F1 0.804 / 0.813 / 0.818 / 0.809 / 0.791. Across all 8{,}040 Gold images, Pearson correlations between MIE probabilities and mean human labels are 0.822 / 0.687 / 0.581. On 186 Gold pairs where both VLM judges contradict the human consensus, MIE follows the humans in 81 cases (43.5\%); MIE therefore does not reproduce every teacher error, although this does not establish that it is more accurate than its teachers.

All training and evaluation are conducted on a single NVIDIA A100 GPU, with 16 vCPUs (Intel(R) Xeon(R) Platinum 8462Y+) and 251 GB of system memory.
\begin{figure}[h]
    \centering
    \includegraphics[width=\linewidth,height=0.6\textheight,keepaspectratio]{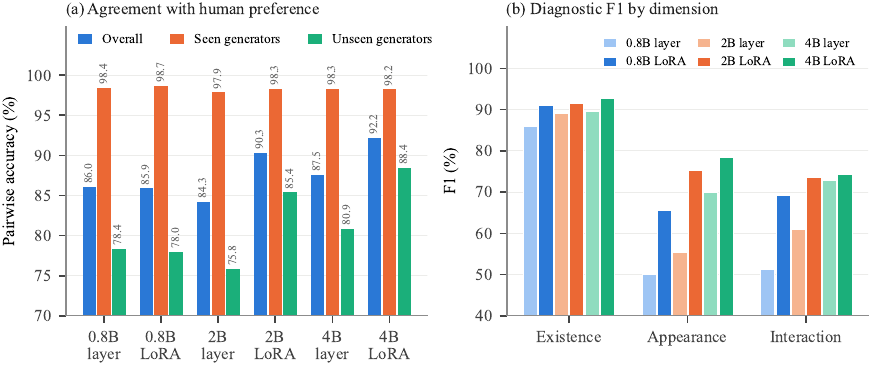}
    \caption{Human alignment of the six MIE variants on the 3{,}690 consensus pairs of MIB-Gold. (a) Overall, seen-generator, and unseen-generator pairwise accuracy. (b) Diagnostic F1 per dimension on images with agreeing labels. The 4B LoRA-layer evaluator is the strongest model on every measure except seen-generator accuracy, where all variants exceed 97.8\%.}
    \label{fig:mie_alignment}
\end{figure}
\begin{table}[h] \centering \caption{LoRA-layer vs.\ layer-only adaptation, evaluated on MIB-Gold. $\Delta$ is the gain of LoRA-layer over layer-only.} \label{tab:lora_ablation} 
\scalebox{0.7}{
\begin{tabular}{@{}lccc@{}} \toprule Backbone & Overall Pairwise Acc. ($\Delta$) & Unseen Split Acc. ($\Delta$) & Diagnostic Macro-F1 ($\Delta$) \\ \midrule 0.8B & $-0.001$ & $-0.004$ & $+0.128$ \\ 2B & $+0.061$ & $+0.096$ & $+0.116$ \\ 4B & $+0.046$ & $+0.075$ & $+0.044$ \\ \bottomrule \end{tabular} 
}
\end{table}

\begin{table}[H]
\centering
\caption{Training Hyperparameters (shared by all six MIE variants, with layer-only variants omitting LoRA; the 0.8B variant is shown).}
\label{tab:training_hparams}
\renewcommand{\arraystretch}{1.1}
\begin{tabular}{l l}
\toprule
\textbf{Parameter} & \textbf{Value} \\
\midrule
Base Model & Qwen3.5-0.8B~\citep{qwen2026qwen35} (via unsloth) \\
Adaptation Method & LoRA~\citep{hu2022lora} ($r=16$, $\alpha=32$, dropout $=0$, no bias) on attention/MLP \\ Additional Unfreezing & Top 4 backbone layers \\
Optimizer & AdamW \\
Learning Rate & $2 \times 10^{-5}$ \\
Weight Decay & 0.01 \\
LR Scheduler & Cosine schedule with warmup \\
Warmup Ratio & 0.03 \\
Batch Size (per device) & 1 \\
Gradient Accumulation Steps & 16 \\
Target Effective Batch Size & 16 \\
Training Budget & 600 optimizer steps ($\approx$9.6K pairs) \\
Gradient Clipping Norm & 1.0 \\
Image Resolution & $512 \times 512$ \\
Global Random Seed & 3407 \\
\bottomrule
\end{tabular}
\end{table}

\begin{table}[H]
\centering
\caption{Multi-task Loss Configuration}
\label{tab:loss_config}
\renewcommand{\arraystretch}{1.1}
\begin{tabular}{l l}
\toprule
\textbf{Component} & \textbf{Setting} \\
\midrule
Preference Ranking Loss & Margin Ranking Loss ($\text{margin} = 0.1$) \\
Ranking Loss Weight ($\alpha$) & 1.0 \\
Diagnostic Classification Loss & BCEWithLogitsLoss \\
Classification Loss Weight ($\beta$) & 0.5 \\
Early Stopping & Patience: 1 epoch, $\text{min\_delta} = 10^{-3}$ \\
\bottomrule
\end{tabular}
\end{table}

\begin{table}[H]
\centering
\caption{Inference and Evaluation Setup}
\label{tab:eval_config}
\renewcommand{\arraystretch}{1.1}
\begin{tabular}{l p{8.5cm}}
\toprule
\textbf{Setting} & \textbf{Detail} \\
\midrule
MIE Inference Precision & Default checkpoint precision (bf16/fp16) \\
Evaluation Resolution & $512 \times 512$ (preserving aspect ratio) \\
Metric 1: Preference Accuracy & Binary choice based on scoring head logits \\
Metric 2: Diagnostic Output & Three per-dimension sigmoid probabilities \\
\bottomrule
\end{tabular}
\end{table}

\begin{table}[H]
\centering
\caption{Baseline Models Configuration}
\label{tab:baselines_config}
\renewcommand{\arraystretch}{1.2}
\begin{tabular}{l p{8.5cm}}
\toprule
\textbf{Model} & \textbf{Configuration} \\
\midrule
CLIP Baseline & \texttt{openai/clip-vit-base-patch32} \newline Text Truncation: Enabled \newline Scoring: Cosine similarity between text embedding and pooled image embedding \\
\addlinespace
DINO Baseline & \texttt{facebook/dinov2-base} \newline Scoring: Cosine similarity between the average normalized subject reference embeddings and generated image embeddings \\
\bottomrule
\end{tabular}
\end{table}

\FloatBarrier
\section{Appendix: Robustness Checks and Breakdown Analysis}
\label{appendix:robust}

\paragraph{Transfer to Unseen References.} We evaluate the unchanged MIE checkpoint on 22 new subjects with zero overlap with MIB's 80 reference subjects: 10 human identities with new demographic and styling combinations (e.g., a Black teen girl with braids, a Latino man in a black polo) and 12 non-human categories absent from MIB (e.g., an alpaca, a desk globe, a violin, a picnic basket), represented by openly licensed real photographs, by synthetic references rendered with the same structured prompts (GPT-Image-1), and by synthetic Qwen-Image references. Table~\ref{tab:unseen_class_auc} reports preference AUC and generator-specific Existence correlations on GPT-Image-1.5 versus Flux.2-klein-9B. The results provide preliminary evidence of transfer on this matchup. \begin{table}[h] \centering \small \caption{Reference-extension results for GPT-Image-1.5 versus Flux.2-klein-9B. AUC uses preference-consistent pairs; Existence correlations use all matched items before preference filtering, reported separately for GPT-Image-1.5 and Flux.2.} \label{tab:unseen_class_auc} \begin{tabular}{@{}lccc@{}} \toprule Reference Source & \# Pairs ($n$) & AUC & Existence $r$ (GPT / Flux) \\ \midrule Real photographs & 173 & 0.827 & 0.932 / 0.575 \\ GPT-Image-1 (synthetic) & 176 & 0.851 & 0.926 / 0.526 \\ Qwen-Image (synthetic) & 181 & 0.775 & 0.938 / 0.476 \\ \bottomrule \end{tabular} \end{table} \paragraph{The Reference-Source Gap Is Generator-Specific and Concentrated in Appearance.} Beyond aggregate AUC, we examine whether real photographic references are intrinsically harder to preserve than synthetic ones. Across the real-photograph and GPT-Image-1 reference sources we collected 864 pairwise comparisons (1,728 ratings from 8 annotators, 2 per pair) covering 1,722 of 1,728 candidate images (six real-reference Flux.2 renders were missing), scored with the same MIE checkpoint reported in the main paper without retraining. Real references are measurably harder, but the effect is generator-specific and concentrated in the appearance facet rather than existence: for Nano Banana, the human-marked appearance-success rate drops from 0.472 with synthetic references to 0.306 with real references, while existence success is essentially unchanged (0.896 real vs.\ 0.889 synthetic). These results do not establish domain invariance. Pooled across the three reference sources, human Appearance agreement is low (Fleiss' $\kappa$ = 0.19 for GPT-Image-1.5 and 0.31 for Flux.2), so conclusions about Appearance alignment remain uncertain. \paragraph{Difficulty of the Tested Novel Subjects.} Table~\ref{tab:longtail_difficulty} compares a per-subject difficulty index (the mean residual image score over scenes containing the subject, where each residual subtracts the mean of its subject-count $\times$ generator $\times$ reference-source cell; pooled over all three reference sources) for scenes containing the 12 new non-human categories and the 10 new human identities. The index is higher for the non-human categories under both human labels and MIE, and MIE's per-subject Existence index correlates with the human index (Spearman $\rho = 0.92$). This comparison does not isolate an effect of category novelty. \begin{table}[h] \centering \small \caption{Per-subject difficulty index (positive: easier than baseline; negative: harder than baseline) for the 22 unseen subjects, split by category.} \label{tab:longtail_difficulty} \begin{tabular}{@{}lccc@{}} \toprule Metric & New Non-Human (12) & New Human Identity (10) & Contrast \\ \midrule Human-marked appearance & +0.0145 & -0.0145 & +0.029 \\ Human-marked existence & +0.0209 & -0.0208 & +0.042 \\ MIE appearance & +0.018 & -0.018 & +0.036 \\ MIE existence & +0.029 & -0.028 & +0.057 \\ \bottomrule \end{tabular} \end{table} \paragraph{MIE Breakdown Analysis} We analyze where the gains of the learned evaluator actually come from. A consistent pattern emerges across all six checkpoints: every model performs better on the seen-generator subset than on the unseen-generator subset, confirming that cross-generator generalization remains a non-trivial challenge. However, the size of this gap depends strongly on the tuning regime. The smallest seen-to-unseen drop is achieved by \texttt{qwen35\_4b\_lora\_layer} at $-0.098$, whereas the largest drop appears in \texttt{2b layer\_only} at $-0.221$. This shows that additional model capacity alone is not enough; how that capacity is adapted is equally important.

The LoRA-versus-layer comparison reinforces this conclusion. At 2B, LoRA-layer tuning improves pairwise accuracy by 0.061 and macro-F1 by 0.116 relative to \texttt{2B layer\_only}. At 4B, it still yields gains of 0.046 in pairwise accuracy and 0.044 in macro-F1 over \texttt{4B layer\_only}. Even at 0.8B, where the pairwise gain is nearly neutral, LoRA-layer tuning improves macro-F1 by 0.128. These results suggest that the primary benefit of LoRA-layer tuning is not merely stronger ranking, but consistently sharper diagnostic discrimination. In other words, the full six-checkpoint comparison supports a benchmark-utility claim rather than a simple ``bigger is better'' story.

\begin{figure}[!ht]
    \centering
    \includegraphics[width=\linewidth,height=0.6\textheight,keepaspectratio]{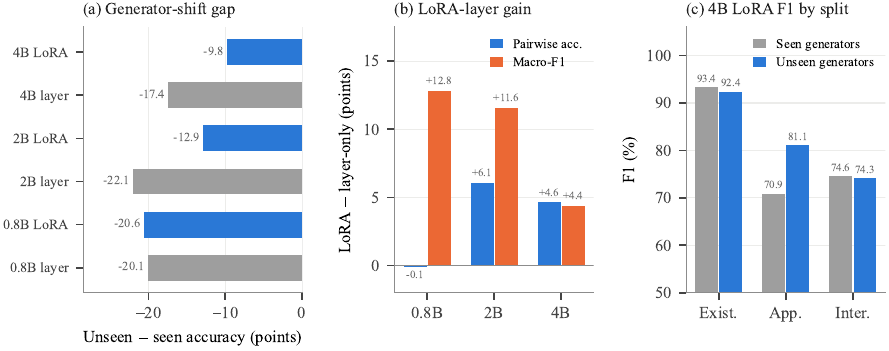}
    \caption{Breakdown of the MIE variants. (a) Unseen-minus-seen pairwise accuracy; the gap depends on both scale and tuning. (b) Gain of LoRA-layer over layer-only tuning: macro-F1 improves at every scale, pairwise accuracy at 2B and 4B but not at 0.8B. (c) Per-dimension F1 of the 4B LoRA-layer model on seen and unseen generators: Existence and Interaction are slightly higher on seen generators, Appearance is higher on unseen generators.}
    \label{fig:mie_breakdown}
\end{figure}

The category-level breakdown points to the same conclusion. \texttt{Existence} remains the easiest dimension overall, whereas \texttt{Appearance} and \texttt{Interaction} provide the more discriminative diagnostic challenge. As shown in Figure~\ref{fig:mie_breakdown}, generator shift is not uniform across dimensions: the strongest checkpoint remains highly stable on \texttt{Existence}, while the more semantic categories continue to differentiate model quality and adaptation strategy. Overall, the breakdown analysis clarifies that the best-performing evaluator is not simply the largest model, but the model that combines additional capacity with the right adaptation strategy and preserves diagnostic sensitivity under distribution shift.

\FloatBarrier
\section{Ethics and Human Annotation}
MIB avoids web-scraped personal photographs: all benchmark reference identities are synthetic, and the real photographs used in the reference-extension study (Appendix~\ref{appendix:robust}) are openly licensed. All human annotations, including the reference-extension study, were performed by the authors themselves, who judged only generated images of synthetic or openly licensed subjects; no external participants were recruited, no compensation was involved, and no personal information about the annotators was collected. Because the annotators were the investigators rather than research subjects, the annotation was not submitted for institutional review board (IRB) review. As authors, all annotators consented to the use of their ratings. Demographic attributes of reference subjects are reported as depicted and were annotated by the authors (Table~\ref{tab:reference_distribution}). Because a stronger evaluator could also be used to optimize personalized generators, including for misuse such as multi-person deepfakes, we release MIE as a diagnostic research tool and discuss dual-use and representational-bias risks in Appendix~\ref{appendix:limit}.

\FloatBarrier
\section{Appendix: Limitations and Societal Impacts}
\label{appendix:limit}
\textbf{Scale and Diversity of the Gold Set.} While our 4K gold evaluation set was curated under a double-blind protocol, it inherently captures a finite subspace of the vast combinations of subjects and interactions. The current benchmark relies on 80 unique reference subjects, which, despite covering major categories of humans, animals, and everyday objects, may not fully represent the long-tail distribution of real-world identities or culturally specific concepts. The white-studio setting and global realistic-scale constraint further narrow the visual distribution to controlled compositional stress, leaving in-the-wild backgrounds and stylistic variations beyond the current scope. Future iterations will expand the reference concept library along long-tail axes and extend evaluation to less constrained visual settings.

\textbf{Temporal Snapshot of Baseline Generators} 
The MIB gold set reflects the performance of six representative generators at the time of collection. 
Given the rapid velocity of the field, newer architectures may encounter unique failure modes, such as specific multi-modal reasoning bottlenecks, not fully captured by the current distribution. 
To ensure the benchmark’s longevity, we commit to releasing versioned extensions as transformative models emerge, providing a standardized re-annotation protocol to facilitate community-driven growth.

\textbf{VLM-Induced Noise in the Silver Set} 
The 60K silver set utilizes a VLM-based pipeline for scalable preference pseudo-labeling. Despite its utility, VLM labeling is susceptible to model-specific biases, such as the tendency to conflate surface-level image fidelity with structural binding accuracy. 
Furthermore, these models may underperform on culturally specific concepts underrepresented in their pretraining data. We recommend that practitioners using the silver set for model alignment (e.g., via DPO) treat these labels as high-fidelity but noisy proxies, incorporating human-in-the-loop auditing for safety-critical applications. 

\textbf{Single-Reference Identity Preservation vs.\ Replication.} A single reference image per subject cannot fully distinguish genuine identity preservation across pose, expression, and lighting variation from trivial replication, the ``copy-paste'' failure mode in which a model directly reproduces the reference face rather than preserving identity under natural variation~\citep{xu2026withanyone}. Two aspects of MIB's design partially mitigate this: Appearance is scored by human/VLM judgment rather than by feature-similarity metrics (e.g., ArcFace, DINO) that would reward exact replication, and Appearance and Interaction are rated as separate facets, so a copy-pasted subject placed in an action-requiring scene is penalized on interaction even though its appearance score stays high. These choices mitigate rather than quantify copy-paste behavior; disentangling genuine preservation from replication requires multiple references per identity across varied poses and conditions, which we leave to future multi-reference extensions of MIB.

\textbf{Positive Impacts.} MIBE contributes a human-aligned diagnostic benchmark (MIB) and evaluator (MIE) that advances transparency and fine-grained error attribution in multi-subject personalized image generation. By decoupling failures into Existence, Appearance, and Interaction rather than reporting a single opaque score, MIBE gives developers actionable signal for reducing hallucinated subjects and identity bleeding in downstream generative systems. \par \textbf{Dataset Impacts.} MIB sets a safer standard for benchmark creation: every released image is either generated synthetically or curated under permissive licenses, which is intended to reduce the privacy, copyright, and harmful-content risks of web-scraped photograph collections; these effects were not measured. \par \textbf{Dual-Use and Deepfake Risks.} A high-fidelity evaluator can also be used to optimize personalized image generators, which could indirectly facilitate more convincing multi-person synthetic media (e.g., deepfakes) if misused. We release MIE as a diagnostic and research tool and encourage downstream users to pair it with existing content-provenance and watermarking safeguards. \par \textbf{Bias and Representation.} MIB's demographic composition, the Silver VLM judges, and MIE's pretrained backbone can carry residual biases from their construction and pretraining. We aim to mitigate this through a gender-balanced human reference pool and dual-VLM consensus annotation, but continued human auditing and further diversification of the benchmark remain necessary before deployment in sensitive real-world applications. 
\end{document}